\documentclass{article} 
\usepackage[preprint]{colm2025_conference}

\usepackage{microtype}
\usepackage{hyperref}
\usepackage{url}
\usepackage{booktabs}
\usepackage{macros}
\usepackage[textsize=tiny]{todonotes}
\usepackage{amsthm}
\usepackage{algorithm}
\usepackage{algorithmic}
\usepackage{booktabs}
\usepackage{mathtools}
\usepackage{listings}
\usepackage{color}
\usepackage{xcolor}
\usepackage{needspace}
\usepackage{hyperref}
\usepackage{url}
\usepackage{makecell}
\usepackage{tcolorbox}
\usepackage{framed}
\usepackage{array}
\usepackage{enumitem}
\usepackage{multirow}
\usepackage{booktabs}
\usepackage{amssymb}
\usepackage{wrapfig}
\newcommand{\rebuttal}[1]{{\color{black}#1}}
\theoremstyle{plain}
\newtheorem{theorem}{Theorem}[section]

\theoremstyle{definition}
\newtheorem{definition}[theorem]{Definition}

\theoremstyle{remark}

\definecolor{lightblue}{RGB}{235,245,255}
\definecolor{lightgreen}{RGB}{220,255,220}
\definecolor{lightyellow}{RGB}{255,255,220}
\definecolor{lightpurple}{RGB}{245,235,255}
\definecolor{softgreen}{RGB}{0,150,0}
\definecolor{softorange}{RGB}{218,165,32}
\newcommand{\entityrel}[1]{\textcolor{blue}{\texttt{$\rightarrow$ (#1)}}}
\newcommand{\correctanswer}[1]{\textcolor{green!50!black}
{\textbf{#1}}}
\newcommand{\componentheader}[2]{{\color{#1}\textbf{#2}}}
\lstdefinestyle{customcode}{
    basicstyle=\ttfamily\small,
    breaklines=true,
    breakatwhitespace=true,
    columns=fullflexible,
    keepspaces=true,
    frame=single,
    framesep=10pt,
    framexleftmargin=5mm,
    backgroundcolor=\color{gray!5},
    showstringspaces=false,
    numbers=none,
    captionpos=b,
    keywordstyle=\bfseries\color{blue!70!black},
    commentstyle=\itshape\color{green!50!black},
    stringstyle=\color{orange!80!black},
    emphstyle=\bfseries\color{red!60!black},
    emph={[2]\bfseries\color{purple!70!black}},
    emphstyle={[2]\bfseries\color{purple!70!black}},
    emph={[2]^question:,^answer:,^explanation:,^Options:},
    moredelim=[is][\bfseries\color{purple!70!black}]{@}{@},
    moredelim=[is][\bfseries\color{red!60!yellow}]{<}{>},
}
\newtcolorbox{grammarbox}[1][]{%
  enhanced,
  title={Knowledge Graph Grammar},
  fonttitle=\bfseries,
  colback=white,
  colframe=black,
  boxrule=0.5pt,
  arc=0pt,
  outer arc=0pt,
  #1
}

\lstdefinestyle{grammarstyle}{
    basicstyle=\ttfamily\small,
    columns=fullflexible,
    keepspaces=true,
    mathescape=true,
    frame=single,
    framesep=3pt,
    numbers=none,
    breaklines=true,
    prebreak=\textbackslash,
    postbreak=,
    breakindent=0pt,
}

\newcounter{kggram}
\renewcommand{\thekggram}{\arabic{kggram}}
\newcounter{kggramsubeq}[kggram]

\newcounter{kggramline}[kggram]

\usepackage{lineno}

\definecolor{darkblue}{rgb}{0, 0, 0.5}
\hypersetup{colorlinks=true, citecolor=darkblue, linkcolor=darkblue, urlcolor=darkblue}

\title{Certifying Knowledge Comprehension in LLMs}

\author{Isha Chaudhary\thanks{Corresponding author. Contact at isha4@illinois.edu}, Vedaant V. Jain \& Gagandeep Singh \\
Siebel School of Computing and Data Science\\
University of Illinois, Urbana-Champaign\\
USA
}

\begin{document}

\ifcolmsubmission
\linenumbers
\fi

\maketitle

\begin{abstract}

Large Language Models (LLMs) are increasingly deployed in safety-critical systems where they provide answers based on in-context information derived from knowledge bases. 
As LLMs are increasingly envisioned as superhuman agents, their proficiency in knowledge comprehension---extracting relevant information and reasoning over it to answer questions, a key facet of human intelligence---becomes crucial.
However, existing evaluations of LLMs on knowledge comprehension are typically conducted on small test sets, but these datasets represent only a tiny fraction of the vast number of possible queries. Simple empirical evaluations on these limited test sets raises concerns about the reliability and generalizability of the results.

In this work, we introduce the first specification and certification framework for knowledge comprehension in LLMs, providing formal probabilistic guarantees for reliability. Instead of a fixed dataset, we design novel specifications that mathematically represent prohibitively large probability distributions of knowledge comprehension prompts with natural noise, using knowledge graphs. From these specifications, we generate quantitative certificates that offer high-confidence, tight bounds on the probability that a given LLM correctly answers any question drawn from the specification distribution.
We apply our framework to certify SOTA LLMs in two domains: precision medicine and general question-answering. Our results reveal previously unrecognized vulnerabilities in SOTA LLMs due to natural noise in the prompts. Additionally, we establish performance hierarchies with formal guarantees among the SOTA LLMs, particularly in the context of precision medicine question-answering.

\end{abstract}

\section{Introduction}

\begin{wrapfigure}[15]{r}{0.4\textwidth}
\centering
\vspace{-5mm}
    \includegraphics[scale=0.2]{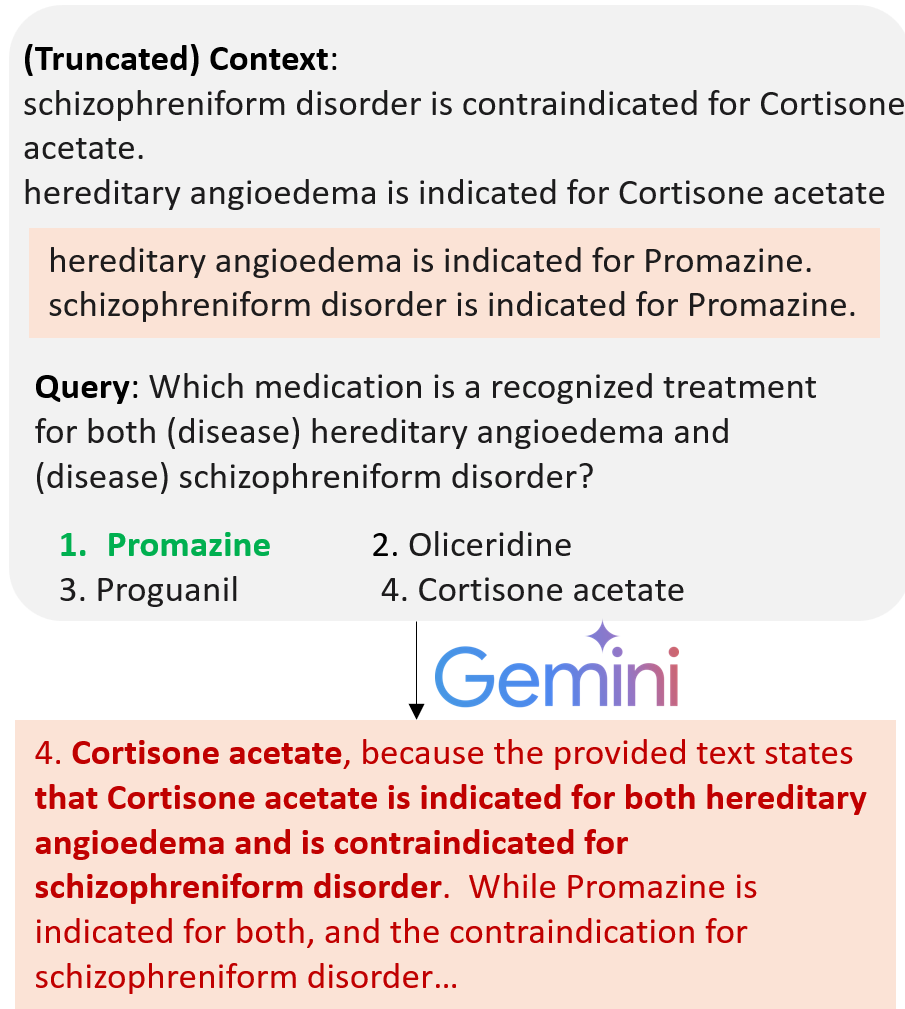}
    \label{fig:motiv}
\end{wrapfigure}


Large Language Models (LLMs) have demonstrated human-level performance in various real-world tasks~\citep{street2024llmsachieveadulthuman,yang2023exploring,opportunitiesofllms,harrison2024comparisonlargelanguagemodel}. A notable application of LLMs is in question-answering systems, including chatbots, which are prompted with relevant, but unstructured information from knowledge bases to generate accurate answers---an ability we refer to as knowledge comprehension.
Knowledge comprehension is a crucial aspect of language understanding, typically assessed in human learners across all education levels~\citep{bloom1956taxonomy, NCES2024ReadingNAEP, ETS2024TOEFLReading, IDPIELTSReading2024}. However, despite being envisioned as superhuman agents~\citep{xi2023risepotentiallargelanguage}, SOTA LLMs often struggle with basic knowledge comprehension tasks in real-world applications. For example, in the adjacent figure, Gemini-1.5-Flash gets confused with the in-context information and provides an incorrect and potentially fatal response in a critical scenario.


Given the widespread relevance of knowledge comprehension tasks, numerous benchmarking studies have been conducted to evaluate LLMs in this domain~\citep{helm, chen2021hybridqa, yang2018hotpotqa, wang2023selfprompted, tang2024multihoprag}. 
%
%
However, their results have been inconsistent. For example, \citet{wei2023largerlanguagemodelsincontext} demonstrates that larger models are robust to label noise in test sets, while \citet{shi2024largerlanguagemodelsincontext} argues that such models can become distracted by this noise. Conversely, \citet{olsson2022incontextlearninginductionheads} indicates that models exhibit similar induction heads for in-context learning, regardless of their size. These discrepancies can be attributed to the following inherent limitations in the nature of these evaluations.


Firstly, LLMs are trained on all available texts, including the standard benchmarking datasets. As a result, models may have encountered the exact prompts used in some test sets, leading to test set leakage during evaluation~\citep{mirzadeh2024gsmsymbolicunderstandinglimitationsmathematical}. Consequently, evaluations relying on these test sets become unreliable.
Secondly, even in cases where the benchmark studies construct their own datasets, these evaluations remain empirical and assess only a limited subset of potential inputs from the vast space of possible knowledge comprehension prompts. 
Thirdly, for deployment in practical scenarios, it is also crucial for the evaluations to be comprehensive and reliable - potentially with a guarantee for their validity.
Therefore, the following question arises: \textbf{How can we assess LLM performance in knowledge comprehension tasks, ensuring reliability of accuracy and generalizability?}


We argue that to reliably deploy LLMs in safety-critical applications, their empirical evaluation on limited test sets is not sufficient. Instead, they must be evaluated on diverse and large sets of prompts (consisting of different querying styles, large in-context information that can distract models like above, noise, etc.) for knowledge comprehension. Further instead of naively creating prohibitively large test sets, we propose to use precise mathematical representations of these prompts, aka \emph{specifications}, to enable a computationally tractable evaluation of LLMs. For reliability of the evaluations, it is desirable to have guarantees on their validity. 
Although prior works~\citep{formalspec4dnn, deeppoly, shi2020robustness, transformer_cert} have proposed formal specifications and certification with guarantees as a method to evaluate traditional neural networks, we cannot directly use them. This is because there exist no formal specifications for knowledge comprehension. Also, certifying LLMs is hard due to a high number of model parameters and specialized nonlinearities, for which traditional certifiers would lose significant precision, leading to inconclusive analysis. 

\textbf{Our approach}. Our work proposes a novel formal specification and certification framework, \tool{} (\textbf{LLM} \textbf{Cert}ification of Knowledge \textbf{C}omprehension) that provides reliable performance assessment of the knowledge comprehension capabilities of LLMs. To encode meaningful and diverse knowledge comprehension prompts, we rely on a structured representation of ground truth knowledge - a knowledge graph. This representation enables us to define new probabilistic specifications, where each specification represents a distribution over knowledge comprehension prompts. The prompt distributions, derived from knowledge graphs, are generated with prohibitively large support, consisting of practical prompts with queries over the knowledge graph. These prompts contain substantial amounts of unstructured, relevant information, along with naturally occurring noise---such as distracting texts and information shuffling---that can influence the LLM's responses. A certificate for a given specification quantifies the probability that the LLM provides a correct response to any prompt sampled from the specification distribution.

\tool{} employs a quantitative certification approach that overcomes the limitations of traditional methods for LLM architectures by using a query-based method relying on input-output examples. This also enables using \tool{} for closed-source, API-accessible models. In contrast to quantitative certification, binary certificates~\citep{ai2, WangZXLJHK:21} for specification compliance can be trivially invalidated due to the ease of constructing failure examples where the desirable property does not hold~\citep{xu2024an, priming}. \tool{} generates probabilistic formal guarantees consisting of high-confidence bounds on the probability of correct knowledge comprehension. \tool{} leverages binomial proportion confidence intervals~\citep{clopper-pearson} to generate high-confidence, tight certification bounds to estimate the underlying probability measure. The use of bounds is particularly advantageous as they also account for the inherent uncertainty in the estimation.
We certify with 2 practical knowledge graphs - PrimeKG~\citep{prime} over precision medicine knowledge and Wikidata5m~\citep{wikidata5m} over general knowledge. We use the generated certificates to establish novel performance hierarchies with formal guarantees among SOTA LLMs. We also show a consistent decline in LLM performance due to noise in prompts.
While formal analysis has been conducted on individual generations of LLMs~\citep{conformallm} and on counterfactual bias~\citep{chaudhary2024quantitativecertificationbiaslarge} in prior work, there is no analysis for the average-case performance of LLMs in knowledge comprehension. Figure~\ref{fig:overview} gives an overview of \tool{}.


\textbf{Main Contributions}: 
\begin{enumerate}
    \item We define knowledge comprehension desirable from LLM responses as a formal specification, using knowledge graphs. We develop novel specifications over large sets of prompts with natural noise, such as distracting text and shuffled information order. We generate and experiment with specifications in safety-critical applications in precision medicine and general question-answering using popular knowledge graphs such as PrimeKG and  Wikidata5m.
    \item We model certification of any target LLM with query-access for a given specification as a probability estimation problem and leverage confidence intervals to generate high-confidence guaranteed bounds on the probability of correct knowledge comprehension. We open-source our code at \url{https://github.com/uiuc-focal-lab/LLMCert-C} and provide a note to practitioners in Appendix~\ref{sec:practitioner}.
    \item We observe that as model parameters increase, knowledge comprehension in LLMs improves with high confidence. We show a consistent decline in SOTA LLMs' performance at answering practical prompts with natural noise.
\end{enumerate}
\section{Certifying knowledge comprehension}

\begin{figure*}[!tb]
    \centering
    \includegraphics[width=0.9\textwidth]{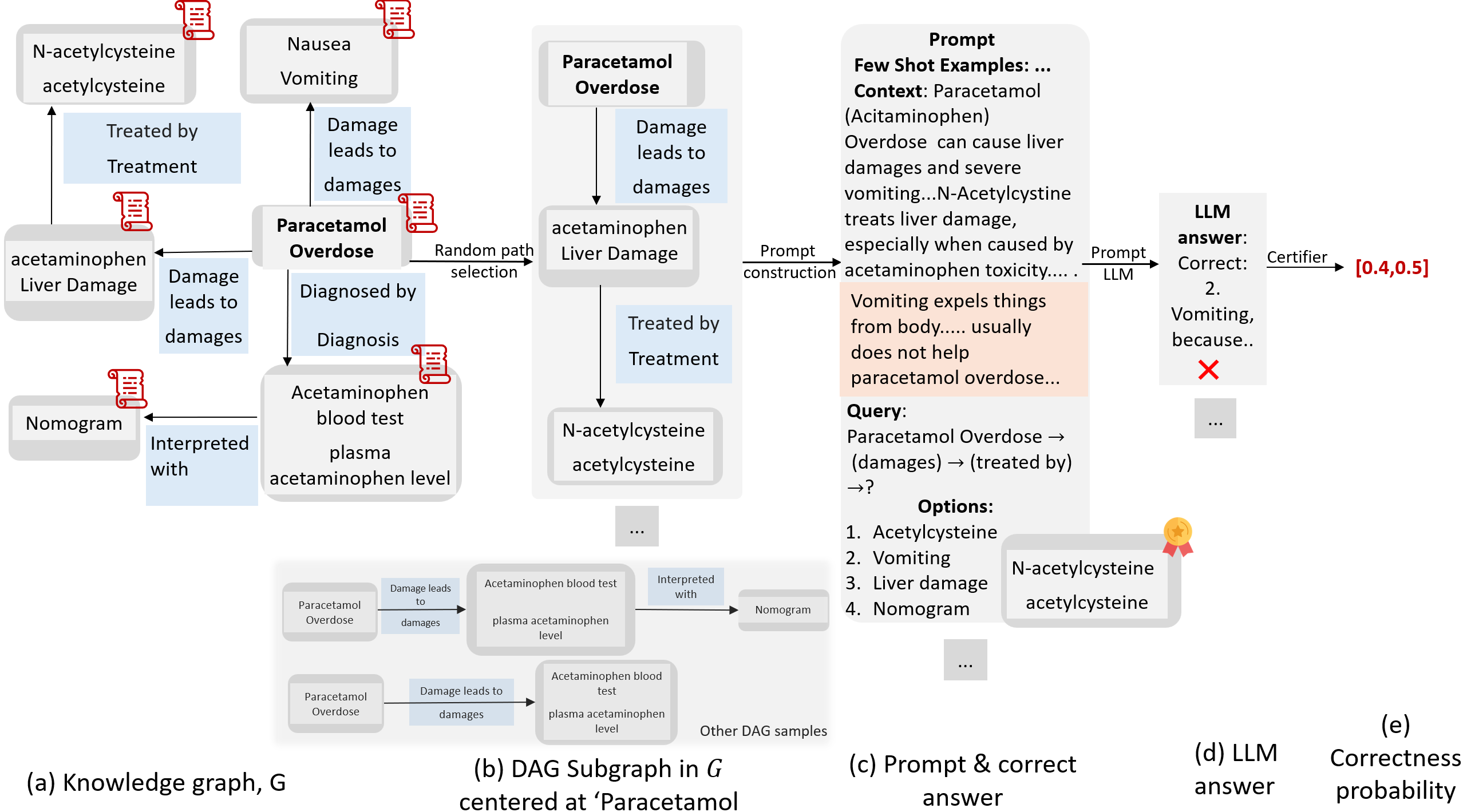}
    \caption{Overview of \tool{}. (a) A KG $\kg$. (b) A randomly chosen path DAG $\kg'$ originating at  fixed node `Paracetamol Overdose', from the various other possible DAGs from node in $\kg$. (c) A prompt created from $\kg'$ having contexts of the nodes in $\kg'$, a distractor context (highlighted in orange, as the node for `Vomiting' is a distractor for $\kg'$), and a query from $\kg'$. (d) The target LLM's response to the prompt, validated using the correct answer. (e) Certifier obtains bounds on the probability of correct response using $\numobs$ samples of LLM responses to randomly sampled prompts.}
    \label{fig:overview}
\end{figure*}

Knowledge comprehension is the ability of a model to retrieve and reason over relevant information from context or parametric knowledge to answer questions succinctly and correctly, according to the provided context.
Figure~\ref{fig:overview} gives an overview of our knowledge comprehension certification framework, \tool{}. Our framework is agnostic to the internal structure of the target model $\llm$ allowing any text-to-text generating model.

\subsection{Specifying knowledge comprehension}

We formally define knowledge comprehension using a knowledge graph (KG, \S\ref{sec:kg}) $\kg = (\nodes, \edges)$.
The adjacent grammar defines a KG. Let $\vocab$ denote the vocabulary of $\llm$'s tokens. $\vocab^+$ denotes the set of concatenation of non-empty sequences of elements of $\vocab$. \begin{wrapfigure}[10]{r}{0.3\textwidth}
\vspace{0.5em}
\refstepcounter{kggram}
  \centering
  \begin{minipage}[c][80pt]{0.3\textwidth}
    \hrule\vspace{0.4em}
    \centering
    KG Grammar
    \vspace{0.2em}\hrule
    \vspace{0.5em}
    \begin{tabular}{@{}r@{\hspace{0.5em}}l @{\hspace{1em}:= \hspace{1em}} l@{}}
        \stepcounter{kggramline}\arabic{kggramline}\refstepcounter{kggramsubeq}.\label{kggram:gamma} & $\gamma$ & $\mathcal{V}^+$ \\
        \stepcounter{kggramline}\arabic{kggramline}\refstepcounter{kggramsubeq}.\label{kggram:eta} & $\eta$ & $\mathcal{V}^+$ \\
        \stepcounter{kggramline}\arabic{kggramline}\refstepcounter{kggramsubeq}.\label{kggram:A} & $\mathcal{A}$ & $[\eta_1, \eta_2, \dots]$ \\
        \stepcounter{kggramline}\arabic{kggramline}\refstepcounter{kggramsubeq}.\label{kggram:node} & $v$ & $(\gamma, \mathcal{A})$ \\
        \stepcounter{kggramline}\arabic{kggramline}\refstepcounter{kggramsubeq}.\label{kggram:edge} & $e$ & $((v_1, v_2), \mathcal{A})$ \\
        \stepcounter{kggramline}\arabic{kggramline}\refstepcounter{kggramsubeq}.\label{kggram:N} & $\mathcal{N}$ & $[v_1, v_2, \dots]$ \\
        \stepcounter{kggramline}\arabic{kggramline}\refstepcounter{kggramsubeq}.\label{kggram:E} & $\mathcal{E}$ & $[e_1, e_2, \dots]$ \\
        \stepcounter{kggramline}\arabic{kggramline}\refstepcounter{kggramsubeq}.\label{kggram:kg} & $\mathcal{G}$ & $(\mathcal{N}, \mathcal{E})$ \\
    \end{tabular}
    \vspace{0.4em}\makebox[\linewidth]{\hrulefill}
  \end{minipage}
  \label{kggram:\thekggram}
\end{wrapfigure}
Let each node $\node$ of $\kg$ (line~\ref{kggram:node}) consist of a finite list of synonymous names (a.k.a. aliases, $\aliases$) that can be used to refer to the node, and a context $\gamma$ that provides more information about the node and its relations with other nodes.

Let $\node_\aliases$ and $\node_\gamma$ denote the aliases and context of $\node$ respectively. For example, in Wikidata5m~\citep{wikidata5m}, each node has aliases consisting of the identifiers mentioned for the subject of the corresponding Wikipedia page and context which is the abstract of the page. Each (directed) edge $\edge$ in $\kg$ (line~\ref{kggram:edge}) is an ordered pair of related nodes where the relation is identified by a set of synonymous aliases $\aliases$ for the edge. Let $(\node_1,\node_2)$ denote any edge from node $\node_1$ to $\node_2$ in $\kg$, where $(\node_1,\node_2)_\aliases$ denotes the aliases of the edge. $\kg$ (line~\ref{kggram:kg}) is a finite collection of nodes $\nodes$ and edges $\edges$. 

A directed acyclic sub-graph (DAG) $\kg'=(\nodes',\edges')$ of $\kg$ is a subgraph of $\kg$ ($\nodes'\subseteq\nodes,\edges'\subseteq\edges$) consisting of directed edges and with no cycles.
Let the nodes of $\kg'$ be topologically ordered as $\mypath'=[\node_1,\node_2\dots,\node_\pathlength]$. Let the $i^{th}$ ($i\in[1,\pathlength]$) nodes of $\mypath'$ from $\node_1$ and backwards from $\node_\pathlength$ be $\mypath'[i]\coloneqq\node_i$ and $\mypath'[-i]\coloneqq\node_{\pathlength-i+1}$ respectively. Nodes $\mypath'_{source}$ with no incoming edges are source nodes of $\kg'$. Nodes $\mypath'_{sink}$ with no outgoing edges are sink nodes of $\kg'$.
%
DAGs generalize paths that are typically used to define multi-hop knowledge comprehension queries~\citep{yang2018hotpotqa,wikihop}. Hence, we define multi-hop knowledge comprehension queries with DAGs in Definition~\ref{def:multihop}.


\begin{definition}
    \label{def:multihop} (Multi-hop queries). Consider a DAG $\kg'$ in $\kg$ with nodes $\mypath'$. A multi-hop query $\query$ for $\mypath'$ is to identify the sink nodes $\mypath'_{sink}$, given aliases of source nodes $\mypath'_{source}$ and aliases of all edges in DAG $\kg'$. To avoid ambiguity regarding correct answer, we enforce $\mypath'_{sink} = \{\mypath'[-1]\}$ in DAGs for multi-hop knowledge comprehension queries.
    %
\end{definition}


The $\kg$ naturally encodes several multi-hop queries, that form the specification distributions for a language model $\llm$. To aid reasoning and reduce dependence on parametric knowledge, prompts include 
 alongside relevant textual context needed to identify the intermediate and final nodes. The context may contain natural noise, such as distractor texts or jumbled information, may occur in practical settings. We model such noise in our prompt distributions as described next.

\subsubsection{Natural noise in prompts}\label{sec:spec_perturbations}
\textbf{\emph{Distractors}}. Prior works~\citep{distractor} on analyzing reasoning in Language Models (LMs) have shown the negative influence of irrelevant information (distractor) in prompts on LM performance. Hence, we include distractor texts in prompt contexts and specify that model response should not use the distractor information. Contexts of nodes $\distnode$ adjacent to any node $\mypath[i]$ ($i\in[1,\pathlength]$) in $\mypath$ that are not the sink node or its predecessors, such that the aliases of the relation $(\mypath[i],\distnode)$ are the same as those of $(\mypath[i],\mypath[j])$, $j > i$ in the DAG, can serve as effective distractors for $\query$ (Definition~\ref{def:distractor}). This is because, at any intermediate step, the LM $\llm$ can pick $\distnode$ as the response, which can deviate $\llm$'s reasoning from $\mypath$. Nodes adjacent to $\mypath[-1]$ and its predecessors in $\mypath$ are not distractors. For the former, the LM must have already reached the final answer before reaching its adjacent nodes, hence answering $\query$. In the latter, adjacent nodes following same relation are valid answers, not distractors.
\begin{definition}
    \label{def:distractor}
    (Distractor). Consider DAG $\kg'$ in $\kg$, with nodes $\mypath'$ and $\mypath$ respectively. A distractor node $\distnode\notin\mypath'$ satisfies $\exists v\in \mypath\setminus\mypath',((v,\distnode)\in\edges)\wedge(\exists \node'\in\mypath',(v,\distnode)_{\aliases}=(v,\node')_{\aliases}\wedge\node'\neq\mypath'[-1])$.
\end{definition} 

\textbf{\emph{Shuffling}}. Prior works~\citep{shuffling} have shown that LM performance can vary with information ordering. Hence, we shuffle the information in provided in prompt, to specify that LM's response should be invariant to information ordering. 


Our property quantifies the probability of observing correct knowledge comprehension for a random multi-hop reasoning prompt, with possible natural noise, developed from $\kg$. We formally define the property for $\llm$ as a probabilistic program over $\kg$ in Alg~\ref{alg:prob_prog_spec}. We follow the syntax of the imperative probabilistic programming language in~\citep[][Figure 3]{prob_prog_syntax}. The language has primitives for sampling from common distributions like Uniform ($\uniform$), Bernoulli ($\bernoulli$), etc., and \texttt{estimateProb(.)} function that outputs the probability of a random variable attaining a certain value. We use a generic identifier $\Dist$ (line 1) for samplers of discrete distributions (`$\dots$' denotes samplers for other discrete distributions). We use \texttt{any}(.) function to denote that at least $1$ of its inputs is true.

A prompt for $\llm$ consists of $\query$ and a context $\contexts$ containing information relevant to answer $\query$ (Alg~\ref{alg:prob_prog_spec}, line 6). $\query$ is developed from a randomly sampled DAG $\kg'$, given by the function \texttt{sampleDAG}. We elaborate on \texttt{sampleDAG}() in \S\ref{sec:local_spec}. $\kg'$ is transformed into a natural language query with a randomly-selected template from a set of prespecified templates, $\tau$. The template samples random aliases of nodes in $\mypath'_{source}$ and relations in $\kg'$ to form a natural language query for $\mypath'_{sink}$. 
$\contexts$ is formed by concatenating ($\odot$) contexts for all nodes in $\mypath'$, followed by optional shuffling. We denote distractor text in $\contexts$ as context of randomly sampled nodes from a distribution $\Dist$ over all distractor nodes of $\mypath'$ in $\nodes$. Our final certificate (line 7) quantifies the probability that $\llm$ generates any alias of $\mypath'_{sink}$, which is correct answer. The certificate depends on the different sampling steps and is specific to $\kg$.

\begin{figure}[tb]
    \centering
    \begin{minipage}{0.55\textwidth} 
        \centering
        \begin{algorithm}[H]
    \textbf{Input:} $\llm, \kg, \tau, \texttt{args}$\\
    \textbf{Output:} $\probcomp$
    \small{\begin{algorithmic}[1]
        \STATE $\Dist\coloneqq\uniform\mid\bernoulli\mid...$ \label{alg:prob_dist}
        \STATE $\kg'\coloneqq \texttt{sampleDAG}(\kg,\texttt{args})$\label{alg:spec_path}
        \STATE $\mypath'\coloneqq\texttt{topologicalOrder}(\kg')$
        \STATE $\query\coloneqq \Dist (\tau)(\kg',\mypath')$\label{alg:spec_query}
        \STATE $\contexts\coloneqq \texttt{shuffle}([\mypath'[0]_{\context}, \dots, \mypath'[-1]_{\context}, (\Dist (\nodes))_{\context}, \dots])$\label{alg:context}
        \STATE $\prompt\coloneqq\contexts\odot\query$\label{alg:spec_prompt}
        \STATE $\probcomp\coloneqq \texttt{estimateProb}(\texttt{any}(\llm(\prompt) == \mypath'[-1]_{\aliases}))$\label{algo:spec_prop}
    \end{algorithmic}}
\caption{General certification}
\label{alg:prob_prog_spec}
\end{algorithm}
    \end{minipage}
    \hfill
    \begin{minipage}{0.44\textwidth} 
        \centering
        \begin{minipage}{\textwidth}
            \centering
            \begin{algorithm}[H]
                \textbf{Input:} {$\kg, \maxpathlen$}; \textbf{Output:} {$\kg'$}
                
                \small{\begin{algorithmic}[1]
                    \STATE $\pathlength\coloneqq\Dist(\{2,\dots,\maxpathlen\})$\label{alg:pathlen}
                    \STATE $\mathcal{B}\coloneqq \texttt{boundedDAGs}(\kg,\mypath_{source},\pathlength)$\label{alg:spathsgen}
                    \STATE $\kg'\coloneqq \Dist(\mathcal{B})$\label{alg:spathsample}
                \end{algorithmic}}
            \caption{Entity centric \texttt{sampleDAG}}
            \label{alg:prob_prog_spec_entity}
            \end{algorithm}
        
        \end{minipage}
        

        \begin{minipage}{\textwidth}
            \centering
            \begin{algorithm}[H]
                \textbf{Input:} $\kg, \kg_{ref}$; \textbf{Output:} $\kg'$
                
                \small{\begin{algorithmic}[1]
                    \STATE $\mathcal{I}\coloneqq \texttt{generateIsomorphisms}(\kg,\kg_{ref})$
                    \STATE $\kg'\coloneqq \Dist(\mathcal{I})$
                \end{algorithmic}}
            \caption{Relations centric \texttt{sampleDAG}}
            \label{alg:prob_prog_spec_relation}
            \end{algorithm}
        \end{minipage}
    \end{minipage}
\end{figure}

\subsubsection{Local specifications}\label{sec:local_spec}
We define individual knowledge comprehension specifications on queries over DAGs that share a common characteristic, thus forming \emph{local specifications}~\citep{formalspec4dnn}. Local specifications are commonly defined for and used in neural network verification~\citep{deeppoly,baluta2021scalable}, due to their tractability to verify and simplicity to interpret. Although there can be various common characteristics to specify DAGs, we define two kinds of local specifications with different common characteristics. \texttt{sampleDAG}() randomly samples DAGs from given $\kg$ having the common characteristics.

\textbf{Entity centric}. Entity-centric question answering~\citep{liu-etal-2023-ask,sciavolino-etal-2021-simple,shavarani2024entityretrievalansweringentitycentric} is an important question-answering paradigm, where the focus is on questions related to particular (real-world) entities. It is relevant in various real-world applications such as topic-specific learning~\citep{topic-learning} and assisted reading~\citep{yu-etal-2020-review} for localizing at information of interest.
We specify knowledge comprehension over entity-centric questions by randomly sampling DAGs starting at a fixed set of nodes $\mypath_{source}$. From a practical standpoint, queries from DAGs with longer paths between source and sink nodes can become meaningless (e.g., {\small Paul Sophus Epstein}$\xrightarrow[]{\text{place of death}}\xrightarrow[]{\text{administrative unit}}\xrightarrow[]{\text{country}}\xrightarrow[]{\text{popular artist}}\xrightarrow[]{\text{genre}}$ \textbf{?}), and thus shorter graphs are considered in popular question-answering datasets such as~\citep{yang2018hotpotqa, trivedi2021musique}. Thus, we upper-bound the size of $\mypath'$ considered in the specification, by a hyperparameter $\maxpathlen\geq 2$. Let \texttt{boundedDAGs} be a function that generates all the DAGs in $\kg$ starting from nodes in $\mypath_{source}$, such that the source and sink nodes have paths of at most $\pathlength\in[2,\maxpathlen]$ nodes. \texttt{sampleDAG} returns a randomly sampled DAG $\kg'$ from $\mathcal{B}$, as described in Alg~\ref{alg:prob_prog_spec_entity}.
 

\textbf{Relations centric}. Question-answering datasets are commonly formed with question templates~\citep{kbqa} each having fixed relations (relation-centric) between variable entities of specific types. For example, ``Which drug is a treatment for \$\{disease\}?", where \$\{disease\} can be substituted with any curable disease to form individual questions. 
We define specifications by fixing the structure and relations of DAG and randomly sampling nodes to obtain the final DAG over which (relation-centric) queries are formed. Alg~\ref{alg:prob_prog_spec_relation} defines the sampling procedure for $\kg'$ to obtain relation-centric DAGs from reference $\kg_{ref}$. 

\begin{definition}
    (Isomorphic DAGs) Let finite DAGs $\kg_1 = (\nodes_1,\edges_1)$ and $\kg_2 = (\nodes_2,\edges_2)$ be isomorphic when there exists a bijection $\phi$ between the nodes of $\kg_1$ and $\kg_2$ such that the following conditions hold.
    (1) $\forall \node_1,\node_2\in \nodes_1, (\node_1,\node_2) \in \edges_1 \Leftrightarrow (\phi(\node_1), \phi(\node_2)) \in \edges_2$; (2) $\forall (\node_1, \node_2) \in \edges_1, (\node_1, \node_2)_\aliases = (\phi(\node_1), \phi(\node_2))_\aliases$. \texttt{generateIsomorphisms}() gives all isomorphisms of $\kg_{ref}$ from $\kg$.
\end{definition}

\subsection{Certification method}

Our algorithm certifies the target LLM $\llm$ by computing an interval $[p_l, p_u]$ bounding probability $p$ (Alg~\ref{alg:prob_prog_spec}, line 7) for a given specification distribution over $\kg$ with high confidence. We model $p$ as the probability of setting the underlying boolean random variable $\randvarcert\triangleq({\texttt{any}(\llm(\prompt) == \mypath'[-1]_{\aliases})})$ to true (success). Thus, $\randvarcert\sim\bernoulli(p)$. Exactly determining $p$ would require enumerating over all possible $\prompt$ which can be developed from any DAG of $\kg$ with any random aliases, resulting in an infeasible number of possible prompts, as shown in Appendix~\ref{app:final_prompt}. Moreover, we want our method to generalize to closed-source LLMs as well, where the internal structures of the models are unknown. Hence, we cannot use any symbolic methods~\citep{mirman2020robustnesscertificationgenerativemodels} to determine $p$. Thus, to scalably certify the black-box target LM $\llm$, we estimate $p$ with provably high-confidence (low error) bounds.
Confidence is defined as the probability of the true value $p$ being within the bounds, i.e., $Pr[p\in[p_l,p_u]]$. To establish formal guarantees, we want our estimation procedure to be such that the actual confidence is at least the user-specified confidence level, $1-\conf$ (i.e., $Pr[p_l\leq p\leq p_u] \geq 1-\conf$), where $\conf>0$ is a small constant. Hence, we use Clopper-Pearson (CP) confidence intervals~\citep{clopper-pearson, cp_ref2}, which is a conservative method known to produce guaranteed high confidence intervals. We show empirical evidence of the bounds' conservativeness in Appendix~\ref{app:validity}.
To compute high-confidence bounds on $p$, we make $\numobs$ independent and identically distributed observations of $\randvarcert$, in which we obtain $k$ successes, $k\in[0,\numobs]$. We generate CP confidence intervals with the $\numobs$ observations to bound $p$ with $1-\conf$ confidence.

    
 

\section{Experiments}\label{sec:expts}

Our experiments were run on 2 A100 GPUs with 40GB VRAM each. We certify the following open-source models --- Llama-3.2-instruct 1B, 3B, 11B~\citep{dubey2024llama3herdmodels}, Phi-3 3B and 14B~\citep{abdin2024Phi-3technicalreporthighly} and their 4-bit quantizations to study the effects of quantization on knowledge comprehension. Among closed-source models with API access, we certify Gemini-1.5~\citep{gemini15} Flash-001 and Pro-002 and GPT-4o-0827~\citep{OpenAI2024GPT4o}.
We use PrimeKG~\citep{prime} and Wikidata5m~\citep{wikidata5m} as our knowledge graphs (KGs) (details in Appendix~\ref{app:preprocessingprime} and Appendix~\ref{app:preprocesswiki5mkg} respectively). PrimeKG is a KG for precision medicine, while Wikidata5m is for general knowledge from Wikipedia. These practical KGs are chosen only for illustration, and our framework generalizes to other KGs. We show relation-centric and entity-centric specifications for PrimeKG and Wikidata5,m respectively.

\textbf{PrimeKG}. We manually define 9 relation-centric specifications based on common clinical reasoning patterns. All specifications are shown in \S \ref{sec: prime_query_gen}. Each specification is defined by:
1. A reference DAG ($\kg_{ref}$);
2. A set of natural language templates (5-7 per specification);
3. Constraints on entity types for isomorphic DAG sampling.
\begin{wrapfigure}[4]{r}{0.3\textwidth}
\centering
    \includegraphics[scale=0.25]{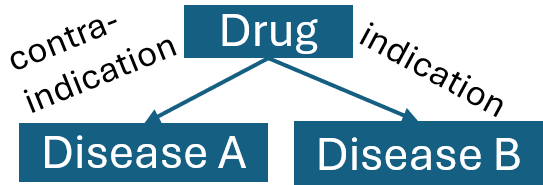}
    \label{fig:motiv}
\end{wrapfigure}An example specification is the ``drug-disease interaction'' specification. It consists of $\kg_{ref}$ which is a 3-node DAG (shown on the right) with drug→disease edges.
An example template to form natural language queries from $\kg_{ref}$ is ``Which drug treats {disease B} but is contraindicated for {disease A}?".

\textbf{Wikidata5m}. 
We created 50 entity-centric specifications using Wikidata5m, sampling linear paths (DAGs) up to length $\maxpathlen=5$ to maintain query relevance. The 50 starting nodes were selected from high out-degree (top 2k) or high-density neighborhood ($>$2k nodes within radius $\maxpathlen$) populations the specifications have a large number of paths, making enumeration infeasible. Note that these specifications are only for demonstration and \tool{} is not limited to them. To avoid bias towards the naturally more frequent shorter paths, we first sample a desired path length uniformly from $[2, \maxpathlen]$. The actual path is then generated from the starting node according to this length (Alg. \ref{alg:prob_prog_spec_entity}).
 We analyze the impact of varying query path lengths on LLM performance in Appendix~\ref{app:ablation}.
Note that the framework is adaptable to other distributions as needed by the specific certification usecase.


\textbf{Specifications with varying complexities}.
We certify $2$ kinds of specifications with different kinds of contexts in LLM prompts (Alg~\ref{alg:prob_prog_spec}, line 4) --- with context shuffling and distractor context (Distractor) and without such noise (Vanilla). This enables us to study the effects of natural noise on LLM performance. 
Leveraging the linear structure of Wikidata5m DAGs, we use weighted sampling (Alg.~\ref{alg:get_best_distractor}) to select distractors closer to the answer node, hypothesizing this increases difficulty. Since PrimeKG lacks such unique linearizations, its distractor nodes are sampled uniformly. We provide an ablation study on the effects of varying the weighted distractor sampling in Appendix~\ref{app:ablation}.
Prompts give multiple answer options with $1$ correct, to simply evaluate LLM responses using string matching (Appendix~\ref{app:checker}).
For answer options in Distractor setting, we prioritize distractor nodes, followed by nodes from the query path, and finally, random nodes related to path nodes. Vanilla setting is similar, without distractors. All further details of prompt construction are in Appendix~\ref{app:promptconst}.

\textbf{Baseline}. We compare \tool{}'s results with a benchmarking baseline. It measures the target LLM's accuracy on a static dataset. We generate 50 static queries from each specification by uniformly sampling DAGs. Queries use a fixed alias of entities and relations, with no shuffling or distractors in contexts to remove natural noise.

\begin{table*}[!tb]
\caption{Average certification bounds for specifications over Wikidata5m and PrimeKG KGs for different LLMs. Bounds are given as [lower bound, upper bound].}
\vspace{0.1cm}
\centering
\scriptsize{
\begin{tabular}{@{} p{2cm} p{1.2cm} ccc ccc @{}}
\toprule
\multirow{2}{*}{Model} & \multirow{2}{*}{Quantization} & \multicolumn{3}{c}{Wikidata} & \multicolumn{3}{c}{PrimeKG} \\
\cmidrule(lr){3-5} \cmidrule(lr){6-8}
& & Baseline & \makecell{Vanilla} & \makecell{Distractor } &  Baseline & \makecell{Vanilla} & \makecell{Distractor} \\
\midrule
{Gemini-1.5-Pro} & - & $0.87$ & {[}0.80, 0.89{]} & {[}0.64, 0.75{]} & $1.00$ & {[}0.97, 1.00{]} & {[}0.92, 0.98{]} \\
 \midrule
{GPT-4o} & -  & $0.88$ & {[}0.80, 0.89{]} & {[}0.62, 0.74{]} & $0.99$& {[}0.97, 1.00{]} & {[}0.85, 0.95{]} \\
 \midrule
{Gemini-1.5-Flash} & - & $0.81$ & {[}0.72, 0.83{]} & {[}0.43, 0.56{]} &  {$0.99$} & {[}0.96, 0.99{]} & {[}0.89, 0.96{]} \\
\midrule
\multirow{2}{*}{Phi-3 (14B)} & fp16 & $0.63$ & {[}0.54, 0.66{]} & {[}0.33, 0.45{]} & $0.98$ & {[}0.93, 0.98{]} & {[}0.75, 0.85{]} \\
 & 4bit & $0.63$ & {[}0.52, 0.65{]} & {[}0.30, 0.42{]} & $0.95$ & {[}0.92, 0.97{]} & {[}0.71, 0.81{]}\\
 \midrule
\multirow{2}{*}{Llama (11B)} & fp16 & $0.61$ & {[}0.50, 0.63{]} & {[}0.33, 0.45{]} & $0.97$ & {[}0.91, 0.97{]} & {[}0.84, 0.92{]} \\
 & 4bit & $0.54$ & {[}0.45, 0.57{]} & {[}0.29, 0.41{]} & $0.94$ & {[}0.88, 0.95{]} & {[}0.79, 0.88{]} \\
 \midrule
\multirow{2}{*}{Phi-3 (3B)} & fp16 & $0.58$ & {[}0.50, 0.61{]} & {[}0.32, 0.45{]} & $0.92$ & {[}0.85, 0.92{]} & {[}0.70, 0.81{]} \\
 & 4bit & $0.57$ & {[}0.48, 0.60{]} & {[}0.29, 0.41{]} & $0.89$ & {[}0.83, 0.91{]} & {[}0.74, 0.83{]}\\
 \midrule
\multirow{2}{*}{Llama (3B)} & fp16 & $0.51$ & {[}0.43, 0.55{]} & {[}0.27, 0.39{]} & $0.87$ & {[}0.80, 0.89{]} & {[}0.70, 0.80{]}\\
 & 4bit & $0.46$ & {[}0.39, 0.51{]} & {[}0.24, 0.35{]} & $0.84$ & {[}0.73, 0.83{]} & {[}0.63, 0.73{]} \\
 \midrule
\multirow{2}{*}{Llama (1B)} & fp16 & $0.38$ & {[}0.30, 0.41{]} & {[}0.22, 0.33{]} & $0.48$ & {[}0.43, 0.56{]} & {[}0.39, 0.52{]} \\
 & 4bit & $0.34$ & {[}0.27, 0.38{]} & {[}0.20, 0.31{]} & $0.38$ & {[}0.33, 0.45{]} & {[}0.28, 0.40{]} \\
\bottomrule
\end{tabular}}
\label{tab:merged_certificates}
\end{table*}

\subsection{Certificates}
\tool{} generates certificates with high confidence ($1-\conf=95\%$), tight lower and upper bounds on the probability of a correct LLM response to a random prompt sampled from the prompt distribution in our specifications. \tool{} uses $\numobs=250$ number of samples. We report the average value of the lower and upper bounds over all specifications over a KG that \tool{} certifies for each LLM, in Table~\ref{tab:merged_certificates}. Each certificate takes 10-20 minutes for generation, where the main bottleneck is each LLM inference. We plan to optimize this in future work.
We also report the average baseline result for each static dataset from the specifications.
Appendix~\ref{app:ablation} presents detailed ablation studies for the impact of the various certification hyperparameters.
Next, we summarize the key observations from Table~\ref{tab:merged_certificates}. 
\begin{figure}[tb]
    \centering
    \includegraphics[width=\linewidth]{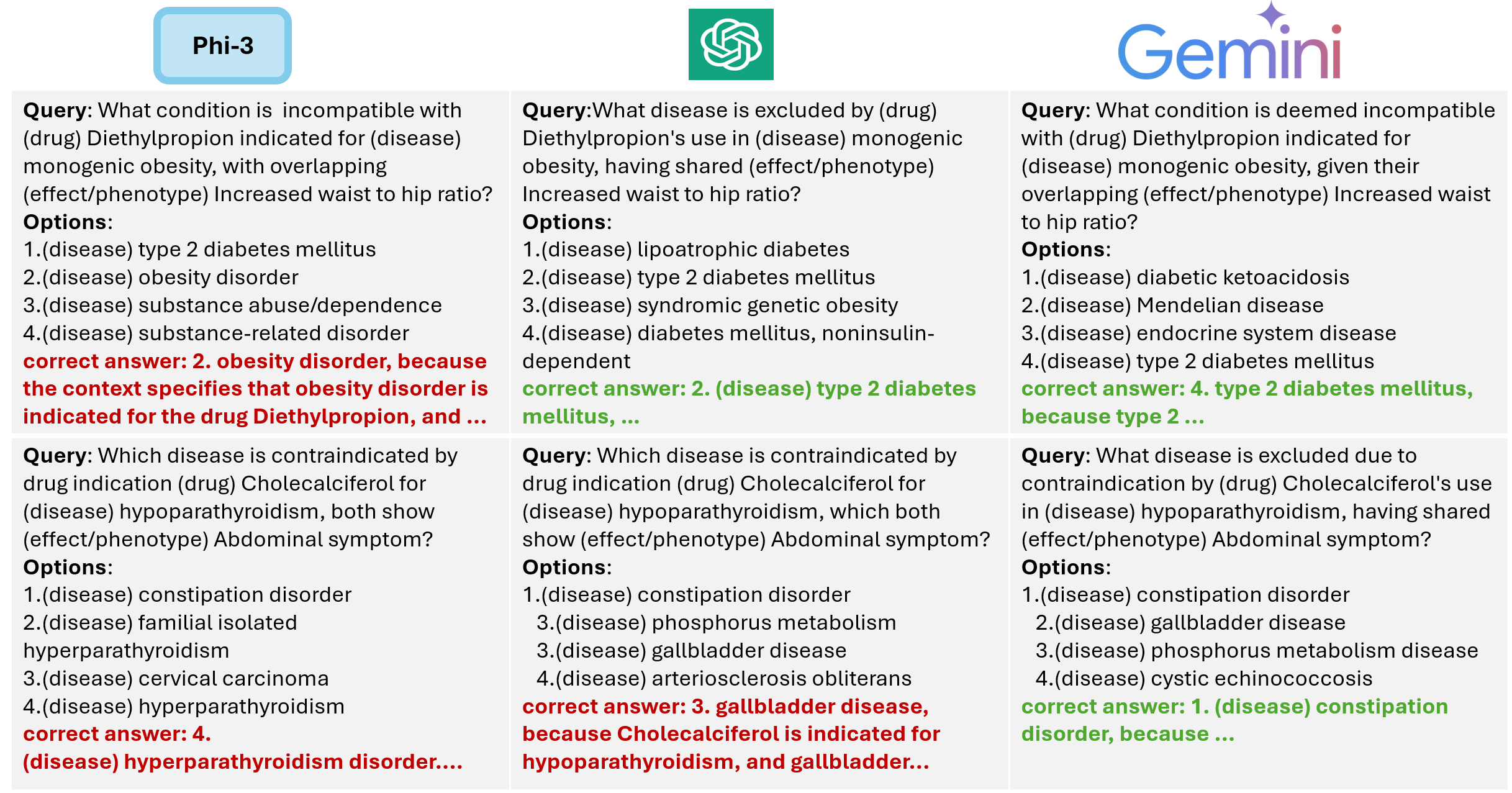}
    \caption{Samples from PrimeKG's Distractor specifications. Prompt context not shown for brevity. Wrong/correct responses are colored red/green respectively. Samples are consistent with certificates, where Phi-3 (14B) has lower bounds than GPT-4o and Gemini-Pro.}
    \label{fig:cert_cs_prime}
\end{figure}

\begin{figure}[tb]
    \centering
    \includegraphics[width=\linewidth]{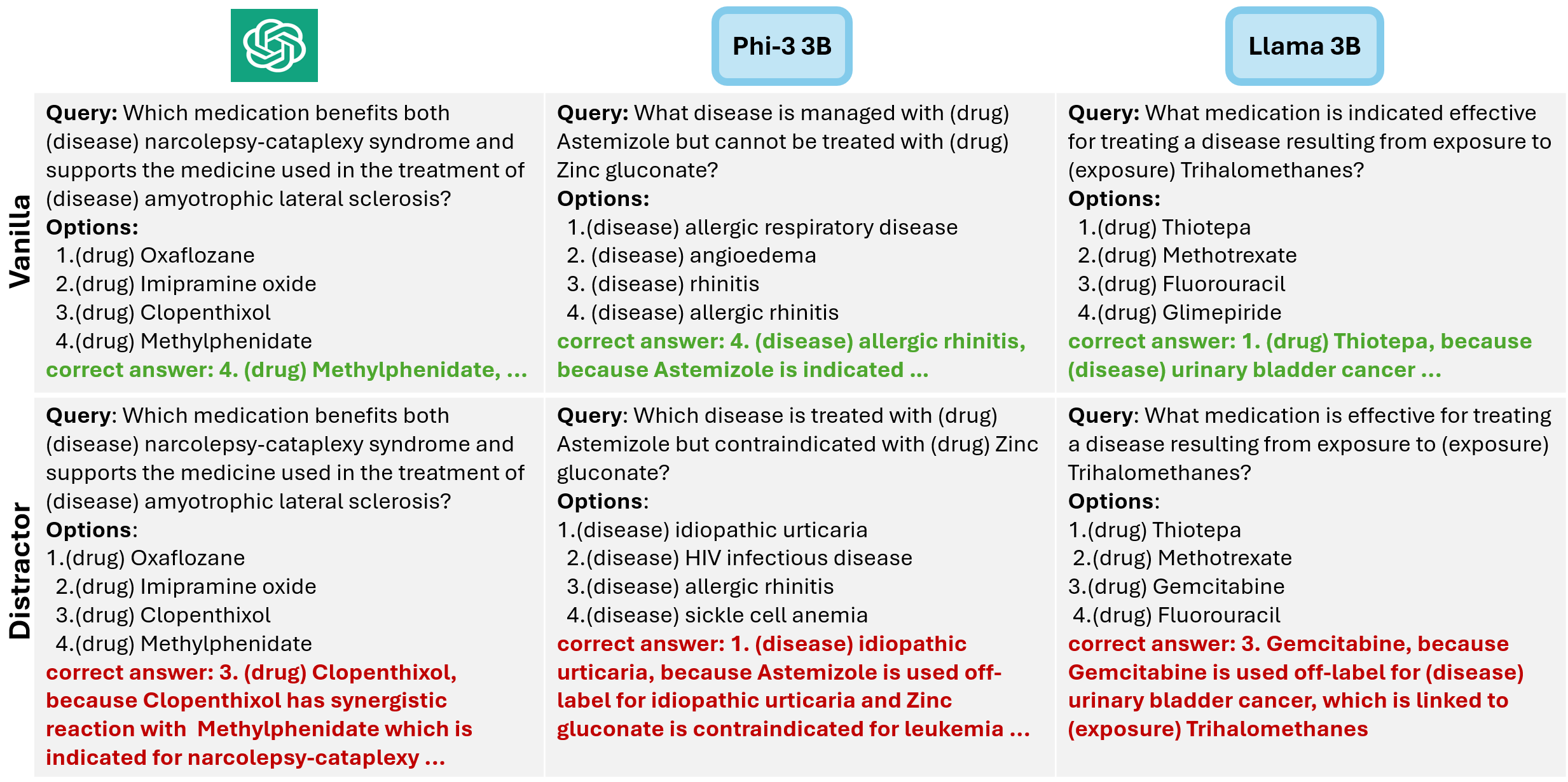}
    \caption{Negative effects of distractors on LLM performance for answering queries from same DAGs (correct and wrong answers in green and red respectively).}
    \label{fig:3_modes_prime}
\end{figure}

\textbf{Variation in knowledge comprehension with models}. We observe that larger models, such as Gemini and GPT, have significantly higher bounds than those for the smaller models, such as Phi-3, Llama, for specifications over both KGs. The lower bounds of the larger models are higher than the upper bounds of the smaller models, suggesting a paradigm shift in knowledge comprehension capabilities (especially for Gemini-1.5-Pro and GPT-4o). However, as the larger models are also closed-source, we are unaware whether the enhanced knowledge comprehension capabilities could be due to specialized training or fine-tuning techniques applied on the models. 
From the certification bounds for PrimeKG, we can establish performance hierarchies with high confidence across the evaluated models. Specifically, when we observe higher lower bound of a model than the upper bound of another model (e.g., Gemini-v1.5-Pro vs Llama 11B for all settings), we conclude the superior performance of the former in the worst case with high confidence. 

\textbf{Effects of different kinds of specifications}. Our results for the different kinds of specifications indicate that Vanilla specifications are easier and lead to higher certification bounds. Distractor specifications are challenging for all models, leading to upper bounds that are consistently lower than the lower bounds for the Vanilla setting. This suggests \emph{susceptibility of SOTA LLMs to natural noise in prompts}. Even larger models such as Gemini-1.5-Pro, which show close to perfect performance in the baseline and Vanilla case (for PrimeKG only), can be incorrect with distractors, showing the nontrivial effects of noise.


\textbf{Comparison with benchmarking baseline}. Baseline scores of all models approach or surpass average certification upper bounds, suggesting inflation of performance estimates in benchmarking. For example, contrary to certification bounds for PrimeKG's distractor specifications, Phi-3 (14B) outperformed Llama (11B). Such findings emphasize the need for more reliable evaluation methods grounded with statistical guarantees. Our certificates also improve upon the baseline by preventing incorrect comparisons between models with similar performance (e.g., Phi-3 14B and Llama 11B and respectively between their 4-bit models for Vanilla specifications) by showing overlaps in their certification bounds.

\textbf{Effects of model quantization}. We see that higher quantization deteriorates model performance on knowledge comprehension, contrary to prior works like \citet{jin2024comprehensiveevaluationquantizationstrategies} that suggest that 4-bit quantization can retain the model's knowledge and reasoning capabilities.





\subsection{Case studies}
Next, we qualitatively analyze certificates for PrimeKG specifications. First, we show the responses of $3$ models in Figure~\ref{fig:cert_cs_prime} --- Phi-3 (3B), GPT-4o, and Gemini-Pro, obtained when certifying them for the Distractor specification. The samples reflect the certification bounds, showing the performance hierarchy across the models. The certification bounds can be used to select the best models for critical applications.
Next, we identify the vulnerabilities of SOTA LLMs to distractor texts in Figure~\ref{fig:3_modes_prime}. We see consistent occurrences of incorrect model responses in the presence of distractors and not otherwise. This demonstrates the vulnerabilities of SOTA LLMs and the risks posed by their deployment in critical applications.
We discuss similar case studies for Wikidata5m specifications in Appendix~\ref{app:cs_wiki}.
\section{Related works}

\textbf{In-context learning}.
As LLM context windows increase~\citep{gemini15,chen2023extendingcontextwindowlarge,dubey2024llama3herdmodels}, more information can be provided in the prompts like few-shot demonstrations~\citep{brown2020languagemodelsfewshotlearners} and examples from related tasks~\citep{qu2024generationalignitnovel}. In-context learning is the emergent behavior~\citep{wei2022emergentabilitieslargelanguage,lu2024emergentabilitieslargelanguage} in which LLMs become proficient at a task with demonstrations in prompts. We use in-context learning and few-shot examples to enhance LLMs' knowledge comprehension.

\textbf{Benchmarking LLM intelligence}.
Several benchmarks have been proposed to study reasoning~\citep{reasoning2,reasoning3,he2024gretrieverretrievalaugmentedgenerationtextual,zha2021inductiverelationpredictionbert}, arithmetic~\citep{arithmetic1,arithmetic2}, planning~\citep{planning1,planning2}, and question-answering~\citep{yang2018hotpotqa,wikihop,kgqa} in LLMs. These benchmarks provide empirical insights and trends of LM performance. However, they use static datasets and not guaranteed to generalize. Certification provides guarantees on the scope (defined by specifications) and confidence of the claims, as shown in this work.



\newpage
\bibliography{colm2025_conference}
\bibliographystyle{colm2025_conference}

\appendix
\newpage



\section{A note for practitioners}\label{sec:practitioner}
Our work takes the first step towards providing formal guarantees of LLMs' knowledge comprehension capabilities, with particular relevance for high-stakes domains like healthcare. While our framework enables more reliable LLM deployment through rigorous guarantees, these certificates should be viewed as probabilistic bounds rather than absolute guarantees of model reliability. We anticipate this work will help make LLMs more trustworthy for safety-critical applications, although continued human supervision remains essential. Future work can extend our framework to certify additional properties like safety and robustness, and integrate it with knowledge graph construction methods to handle less structured inputs. With a one-time effort to construct domain-specific knowledge graphs, practitioners can obtain reliable insights into the performance of LLMs for question-answering and knowledge comprehension tasks. We release our implementation to support further research on formal verification of language models.
\section{Background}
\subsection{Knowledge graph}\label{sec:kg}
\begin{figure}[H]
\centering
    \includegraphics[width=0.25\textwidth]{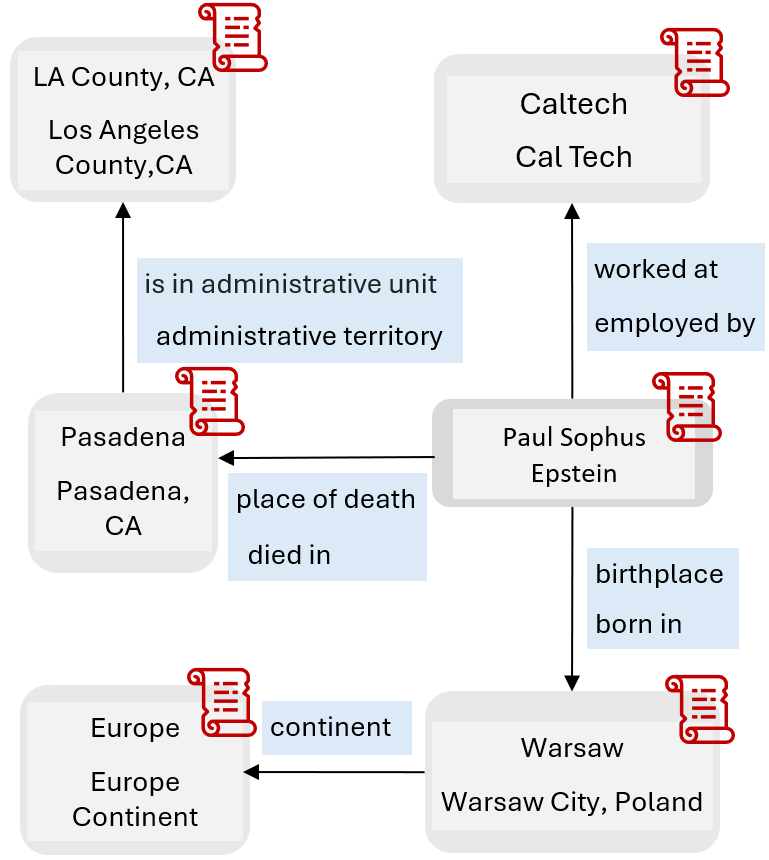}
    \caption{A subgraph of Wikidata5m}
    \label{fig:wikidata_subgraph}
\end{figure}

A knowledge graph $\kg = (\nodes,\edges)$ is a collection of nodes $\nodes$ representing entities, interconnected by directed edges $\edges$ representing their relations~\citep{kg1,kg2}. They are commonly used in search engines to improve the accuracy of responses to user queries. Hence, big companies develop their own closed-source knowledge graphs. PrimeKG~\citep{prime} is an open-source knowledge graph of precision medicine consisting of nodes corresponding to diseases, drugs, effects, etc. PrimeKG describes $\sim$17k diseases with $\sim$4M relations. Wikidata5m~\citep{wikidata5m}, another popular open-source knowledge graph consisting of 5M nodes, is a structured representation of Wikipedia. Each Wikidata node corresponds to a Wikipedia page, containing its abstract and a set of aliases that can synonymously refer to the node. Two nodes $(\node_1, \node_2)$, $\node_1,\node_2\in\nodes$ are connected by a labeled edge if there is a link in the supporting document for $\node_1$ to that for $\node_2$. Edge $(\node_1, \node_2)$ is labeled by a set of aliases for the relation between $\node_1$ and $\node_2$. App. Figure~\ref{fig:wikidata_subgraph} shows a subgraph of Wikidata5m.  

\subsection{Large Language Models}
Large Language Models (LLMs) are autoregressive causal language models that operate on a vocabulary $\vocab$, a set of tokens. LLMs takes a sequence of tokens $x_1, ..., x_n$ where $x_i \in \vocab, n > 0$ and outputs a probability distribution over $\vocab$ to sample the next token $x_{n+1}$. These models are pretrained on vast corpora~\citep{liu2024datasetslargelanguagemodels} and have shown remarkable capabilities ~\citep{llama1, gemini15, OpenAI2024GPT4o}. Numerous benchmarks~\citep{yang2018hotpotqa, rein2023gpqagraduatelevelgoogleproofqa, hendrycks2021measuring} have been developed to evaluate the performance of LLMs on tasks related to multi-step reasoning, knowledge comprehension and question answering. However, there remains a gap in our theoretical understanding of LLMs' capabilities.

\section{Validity of Certification bounds}\label{app:validity}
\begin{figure}[h]
    \centering
    \includegraphics[width=0.5\linewidth]{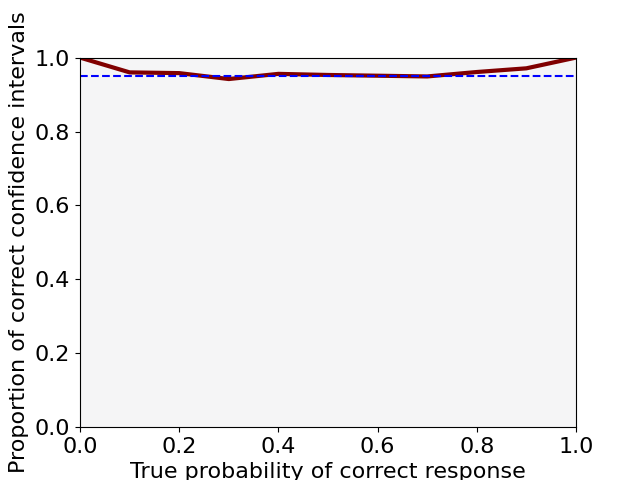}
    \caption{Variation in the proportion of Clopper-Pearson confidence intervals containing the true probability of correct response}
    \label{fig:validity}
\end{figure}

In this section, we provide empirical evidence for the validity of the certification bounds with the user-defined confidence (95\%). We select $10$ equally-spaced values of the true probability of correct LLM response, between 0 and 1. For each true probability, we generate $1000$ Clopper-Pearson confidence intervals, which are used to form our certification bounds. We calculate the proportion of the confidence intervals that contain the true probability and compare with the confidence level. We show in Figure~\ref{fig:validity} that the proportion of instances where the confidence intervals contain the true probability is almost always more than $95\%$.

\section{Knowledge Graph and Query Generation}
This section details our experimental setup for generating multi-hop reasoning queries using the Wikidata5m knowledge graph. We describe the structure of the knowledge graph, the process of generating random paths, formulating queries, and creating answer options including distractors.

\subsection{Knowledge Graph Structure}
Our experiments utilize two knowledge graphs: Wikidata5m and PrimeKG, each with distinct characteristics.
\subsection{Wikidata5m}
Wikidata5m is characterized by rich textual and semantic features:
\begin{itemize}
\item \textbf{Nodes}: Each node represents an entity and is accompanied by a descriptive text paragraph that provides contextual information about the entity.
\item \textbf{Edges}: Relationships between entities are represented as edges.
\item \textbf{Text Paragraphs}: Each node contains an associated text paragraph that that often provides at minimum information relevant to its relationships with other entities. 
\item \textbf{Aliases}: Each node and each edge has a set of aliases associated with them which are just different names for them.
\end{itemize}

\subsubsection{Preprocessing the wikidata5m knowledge graph}~\label{app:preprocesswiki5mkg}
To ensure the generation of unambiguous queries and support the certification process, we preprocess the wikidata5m dataset. 

\begin{enumerate}
\item \textbf{Relation Filtering:} We remove relations such as 'instance of', 'subclass of', and 'part of' due to their inherent potential for ambiguity in query formulation.

\item \textbf{Relevant Information Extraction for edges:} To ensure the relevance of relationships in the knowledge graph, we require textual evidence for each edge. When entity A is related to entity B, we identify specific sentences in the descriptive text of either entity that explicitly mention any alias of the other entity. We assume these sentences support the relationship's existence. These sentences are then linked to the edge, providing context that can be used to answer queries about the relationship. This approach ensures that the knowledge graph contains valid relationships and the specific text that justifies them, enhancing the available context for further analysis. If we find no supporting text for an edge, we drop that edge from the knowledge graph.

\item \textbf{Unicode to ASCII:}For consistency within our experiments,  we convert all text containing Unicode characters into their respective ASCII approximations.
\end{enumerate}

\subsection{PrimeKG}
PrimeKG presents a more streamlined structure:
\begin{itemize}
\item \textbf{Nodes}: The graph consists of nodes representing entities, but unlike Wikidata5m, nodes do not contain accompanying descriptive text.
\item \textbf{Edges}: Similar to Wikidata5m, edges denote relationships between entities.
\item \textbf{Aliases}: Each node and edge is associated with exactly one alias, providing a single alternative name rather than multiple variants.
\end{itemize}

\subsubsection{Preprocessing the PrimeKG knowledge graph}\label{app:preprocessingprime}
The PrimeKG dataset requires several key transformations to align with our experimental framework.

\begin{enumerate}
\item \textbf{Converting to a Directed Graph:} We convert PrimeKG's undirected structure to a directed graph by creating bidirectional edges. Each original edge generates two directed edges with labels incorporating the target entity's type (e.g., an undirected indication edge between a drug and disease yields "indication (drug)" and "indication (disease)" directed edges).

\item \textbf{Enriching Entity Representations:}We augment entity representations by prepending type information to each entity's alias, maintaining a single alias per node to preserve medical precision.

\item \textbf{Generating Supporting Text:} To address PrimeKG's lack of contextual information, we generate descriptive text for each edge using templates tailored to the entity types and relationships (e.g., "entity\_1 is {relation name} for entity\_2"). Implementation details are available in the accompanying code.

\item \textbf{Standardizing Character Encoding:} All text is converted to ASCII format, matching the Wikidata5m preprocessing pipeline.
\end{enumerate}

\subsection{Query Generation}~\label{sec:querygen}

Query generation follows a two-phase process: (1) DAG generation according to the knowledge graph structure and specifications, and (2) transformation of the DAG into a natural language query. The specifics differ between Wikidata5m and PrimeKG due to their distinct structures and certification requirements.

\subsubsection{Wikidata5m}
For Wikidata5m, we generate linear DAGs (paths) from specified pivot nodes. This approach enables focused testing of an LLM's knowledge comprehension capabilities through a chain of relationships. In this case therefore $\mypath_{source}$ has only one node the pivot node, $\node_0$.
Then we simulate the  algorithm \ref{alg:prob_prog_spec_entity} and sample a path for our query generation.

From the pivot node $\node_0$, we construct a local subgraph $\kg(\node_0)$ containing all paths $\mypath_{\node_0}$ originating from $\node_0$. This local subgraph construction serves two purposes: efficient path sampling and enforcement of the $\maxpathlen$ constraint from Algorithm~\ref{alg:prob_prog_spec}. We set $\maxpathlen=5$ as longer paths often lose semantic coherence in knowledge comprehension queries.

Within $\kg(\node_0)$, we generate paths through a randomized sampling process:

1. Sample path length $k_{choice}$ uniformly from $\{1,\ldots,\maxpathlen\}$, following line~\ref{alg:pathlen} in Algorithm~\ref{alg:prob_prog_spec_entity}

2. Generate path through iterative node sampling:
   - Start with $\mypath = [\node_0]$
   - At each step $i$, sample next node according to distribution $\Dist$ (defined in line~\ref{alg:prob_dist}):
     $$\node_{i+1} \sim \Dist([\node' \mid (\node_i,\node')\in\edges \land \node'\in\kg(\node_0)])$$
   where $\node_i$ is the current terminal node of $\mypath$

To ensure well-defined queries per Definition~\ref{def:multihop}, we require path uniqueness: following the relation sequence of $\mypath$ from any node must lead to at most one target node. This constraint ensures $|\mypath'_{sink}| = 1$ as specified in Definition~\ref{def:multihop}, while preserving the natural graph structure where nodes may have multiple edges with the same relation to non-path nodes. Additionally, this approach prevents queries with multiple valid answers, which would complicate the evaluation of the language model's performance.
It's important to note that this uniqueness constraint applies only to the specific path being generated. Nodes within the path may still have multiple edges with the same relation type to other entities not on the path. This allowance maintains the natural complexity of the knowledge graph structure, where entities can have multiple relationships of the same type with different entities.

The complete path generation process is detailed in Algorithm~\ref{alg:pathgen}, which implements the path sampling component of \texttt{sampleDAG} used in lines~\ref{alg:spathsgen}-\ref{alg:spathsample} of Algorithm~\ref{alg:prob_prog_spec_entity}.

The pseudocode for the path generation algorithm is specified in ~\ref{alg:pathgen}

\begin{algorithm}
\caption{Random Path Generation}
\label{alg:pathgen}
\begin{algorithmic}[1]
\STATE \textbf{Input:} Graph $G$, Integer $k$, Vertex $source$
\STATE \textbf{Output:} $path$
\STATE $path\_len \gets$ \textbf{RandomInteger}(1, $k$)
\STATE $path \gets$ None
\WHILE{$path$ is None}
    \STATE $path \gets$ \textbf{DFSPath}($G$, $source$, $path\_len$)
    \IF{not \textbf{IsUnique}($path$)}
            \STATE $path \gets$ None
    \ENDIF
\ENDWHILE
\STATE Return $path$
\end{algorithmic}
\end{algorithm}

\paragraph{Query Formulation: }Once a valid path $\mypath$ is generated, we convert it into a query string. This process aligns with line 3 in Algorithm~\ref{alg:prob_prog_spec}. The query is constructed by sampling aliases for each node and relation in the path. For example, a path $\mypath = [A, B, C]$ might be converted to a query "sampled\_alias(A) $\rightarrow$ sampled\_alias((A, B)) $\rightarrow$ sampled\_alias(B) $\rightarrow$ sampled\_alias((B, C)) $\rightarrow$ ?". Here the tuple of two nodes represents their edge. The aliases are sampled randomly from a discrete uniform distribution over the available aliases for a node or an edge.

\begin{figure}
    \centering
    \includegraphics[width=0.5\linewidth]{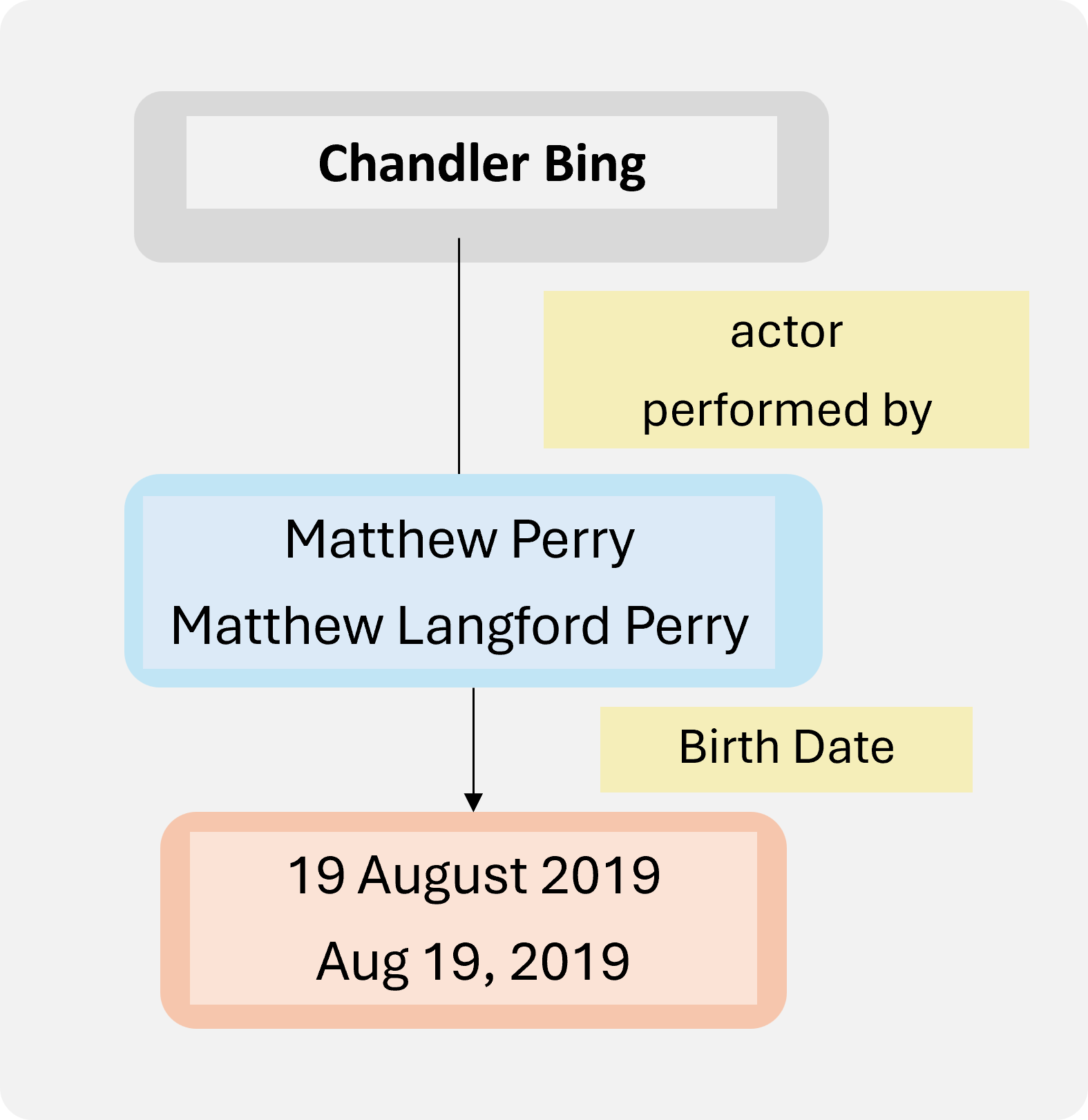}
    \caption{Potential Path in a Subgraph where Pivot is Chandler Bing}
    \label{fig:chandler_path}
\end{figure}

\paragraph{Example Query Generation}
To illustrate our query generation process, consider the scenario of a path in our subgraph as shown in ~\ref{fig:chandler_path}.

Our algorithm would construct the following query from the path presented in ~\ref{fig:chandler_path}:

``Chandler Bing$\rightarrow$(actor)$\rightarrow$(birth\_date)$\rightarrow$?"

This query requires the LLM to reason through two hops in the knowledge graph:
\begin{enumerate}
    \item Identify the actor who played Chandler Bing (Matthew Perry)
    \item Find the birth date of Matthew Perry (19 August 1969)
\end{enumerate}

This example demonstrates how our query generation process creates questions that require multi-hop reasoning, leveraging the structure and relationships within the knowledge graph.

\subsubsection{PrimeKG}
\label{sec: prime_query_gen}
For PrimeKG, each certificate is defined by a reference DAG $\kg_{ref} = (\nodes_{ref}, \edges_{ref})$ that encodes a specific clinical reasoning pattern. Following Algorithm~\ref{alg:prob_prog_spec_relation}, we generate isomorphic DAGs $\kg'$ that preserve the structure and relation types of $\kg_{ref}$ while varying the specific entities.

Formally, for a given $\kg_{ref}$, we sample DAGs $\kg'$ such that there exists a bijection $\phi: \nodes_{ref} \rightarrow \nodes'$ preserving:
1. Edge structure: $(\node_1, \node_2) \in \edges_{ref} \iff (\phi(\node_1), \phi(\node_2)) \in \edges'$
2. Relation types: $(\node_1, \node_2)_{\aliases} = (\phi(\node_1), \phi(\node_2))_{\aliases}$

We define nine core reference patterns capturing common clinical reasoning chains:

\begin{itemize}
    \item Which drug is a contraindication for \#disease A\# and indication for \#disease B\#, when patient has both diseases?
    \item Which drug can be used as indication for <disease A> and is atleast an off-label use drug for \#disease B\#?
    \item Which drug can be used as an indication for \#disease A\# and is an indication for \#disease B\#?
    \item Which drug is an indication for \#disease A\# and interacts synergistically with the treatment of \#disease B\#?
    \item Which drug targets \#genes/proteins(RIDR33)\# associated with \#disease(RIDR38)\#?
    \item Which disease is contraindicated by drug indication \#drug A\# for \#disease A\#, which both show \#effect/phenotype A\#?
    \item Which disease is treated with \#drug A\# but contraindicated with \#drug B\#?
    \item Which drug can be used to treat a disease caused by exposure of \#exposure A\#?
    \item Which disease can be a diagnosis for the following phenotypes - \#list of phenotypes\#?
\end{itemize}

For each reference DAG $\kg_{ref}$, we implement specialized sampling functions that generate isomorphic DAGs while respecting PrimeKG's medical domain constraints. These functions implement the \texttt{generateIsomorphisms} procedure from Algorithm~\ref{alg:prob_prog_spec_relation} as follows:

1. Type constraints: Each node $v \in \nodes_{ref}$ specifies required entity types (e.g., DRUG, DISEASE)
2. Relation constraints: Each edge $e \in \edges_{ref}$ requires specific medical relationships (e.g., indication, contraindication)
3. Clinical validity: Additional constraints ensure sampled DAGs represent valid clinical scenarios

For implementation details of these specialized sampling functions, see the discovery\_functions in our codebase.

\paragraph{Query Formulation: } For a sampled DAG $\kg'$, we generate natural language queries using predefined templates $\tau$ (line~\ref{alg:spec_query} in Algorithm~\ref{alg:prob_prog_spec}). Each certificate has a minimum of 5 semantically equivalent templates that vary in:

1. Syntactic structure: "Which drug treats X and Y?" vs. "Find a medication indicated for both X and Y"
2. Clinical terminology: "contraindication" vs. "should not be used with"
3. Question framing: Direct vs. scenario-based queries


\subsection{Prompt Construction}~\label{app:promptconst}
The final prompt is constructed using a template applied to the query. This process involves several steps, each addressing specific requirements:
\begin{itemize}
    \item Query Formulation: Convert the generated path into a query string as described earlier.
\item Context: This is the supporting text we provide the LLM to answer the query correctly. We additionally trim the context to fit within the LLM's context length limits.
\item Few-shot Examples: Include examples to guide the LLM in understanding the query format and expected answer structure.
\item Answer Options Generation: Create a set of possible answers, including the correct one. The LLM has to choose one of these options as the correct one.
\item Distractors: In the distractor setting, we need to find distractors for the query which need to be included in the prompt.
\end{itemize}
These inclusions ensure that the prompt is comprehensive, fits within model constraints, and provides sufficient guidance for the LLM to generate accurate responses. We also provide information on the aliases used and the entities they correspond to in the prompt, to ensure that the LLM knows about the alias.

\subsubsection{Distractor Selection}
Following Definition~\ref{def:distractor}, we implement distinct distractor selection strategies for each knowledge graph structure to create challenging evaluation scenarios.

\paragraph{Wikidata5m Linear Paths}
For a path $\mypath = [\node_1, ..., \node_k]$, we select distractors through weighted sampling based on path position:

1. \textbf{Distractor Identification}: For each node $\node_i \in \mypath$:
  - Find candidate distractors $D_i = \{\distnode \mid (\node_i,\distnode) \in \edges \land \distnode \notin \mypath\}$
  - Filter to preserve relation type: $D_i' = \{\distnode \in D_i \mid \exists j>i: (\node_i,\distnode)_{\aliases} = (\node_i,\node_j)_{\aliases}\}$

2. \textbf{Position-based Weighting}: Assign weights to $\distnode \in D_i'$:
  $$w(\distnode) = i/|\mypath|$$
  This weights distractors near path end higher, as formalized in Algorithm~\ref{alg:get_best_distractor}.

\paragraph{PrimeKG DAG Structure}
For reference DAG $\kg_{ref}$ and its isomorphic instance $\kg'$, we:

1. Include all valid distractors in context:
  $$D = \{\distnode \mid \exists v \in \nodes': (\distnode \text{ satisfies Definition~\ref{def:distractor}})\}$$

2. Ensure answer options contain at least one distractor:
  $$\text{options} = \{\mypath'_{sink}\} \cup \{d \sim \text{Uniform}(D)\} \cup \text{Others}$$

  This flexible approach demonstrates how our framework is easily adaptable to different kinds of queries.

\subsubsection{Answer Options}~\label{sec:answer_options}

After formulating the query, we generate a set of answer options. This set includes:

\begin{itemize}
    \item The correct answer: The last entity in the generated path.
\item Other entities in the path.
\item Related entities: Entities that share some edge with nodes in the path but are not part of the path.
\item Distractors: A distractor is a node in the knowledge graph $\kg$ that shares a relation with a node in the path, mirroring the relation that continues the path, but the distractor is not itself part of the path. For a formal definition, refer to Definition~\ref{def:distractor}. These are only included in the options in the distractor setting.
\end{itemize}

The process of generating answer options is detailed in Algorithm~\ref{alg:gen_answer_options}. In the algorithm, we sample answer options from the set described above. The answer option algorithm assumes that distractors are input in a list according to the order of preference. Another important note is that for primeKG, we try to ensure that all options are of the same type as the correct answer to prevent the LLM from easily eliminating wrong options.

\begin{algorithm}
\caption{Generate Answer Options}
\label{alg:gen_answer_options}
\begin{algorithmic}[1]
\STATE \textbf{Input:} $correct\_ans$, $distractors$, $path\_entities$, $random\_entities$, $Graph$, $min\_num\_options$
\STATE \textbf{Output:} $options$

\STATE $options \gets$ [($correct\_ans$] $\cup$ $distractors$

\STATE Add path entities to $options$

\STATE Add random entities from $random\_entities$ to $options$

\STATE Return \textbf{Shuffle}($options[:min\_num\_options]$)
\end{algorithmic}
\end{algorithm}

\begin{algorithm}
\caption{Get Best Distractor}
\label{alg:get_best_distractor}
\begin{algorithmic}[1]
\STATE \textbf{Input:} Graph $G$, Path $\mypath$
\STATE \textbf{Output:} $best\_distractor$

\STATE $D \gets$ [] \COMMENT{List of distractors}
\STATE $W \gets$ [] \COMMENT{Weights for distractors}

\FOR{$i \gets$ $0$ to len($\mypath$) $- 2$}
    \STATE $\node \gets \mypath[i]$
    \STATE $N \gets$ GetNeighbors($G$, $\node$)
    \STATE $N\_distractors \gets$ FilterDistractors($N$, $\node$, $\mypath$)
\STATE Extend $D$ with $N\_distractors$
\STATE Extend $W$ with $[i+1] * len(N\_distractors)$
\ENDFOR

\IF{$D$ is not empty}
    \STATE Return \textbf{WeightedRandomChoice}($D$, $W$)
\ELSE
    \STATE Return None
\ENDIF
\end{algorithmic}
\end{algorithm}

\subsection{Context Trimming}~\label{sec:context_trim}
To address the input length limitations of various LLMs, we implement a context trimming procedure. Including all text associated with each node in a reasoning path can result in excessively long contexts. Our procedure aims to preserve the most relevant information from the knowledge graph and supporting texts while respecting each model's maximum input length. This involves identifying relevant sentences per edge in the graph and then trimming the context for each query based on this information.

\subsubsection{Finding Relevant Sentences per Edge (Wikidata5m)}

Each node in the Wikidata5m knowledge graph has associated textual support for its relations. We utilize this textual information to provide query-relevant context. We need to determine the relevant information from the textual supports for each edge as this would help us trim the contexts accordingly.
For each edge $(u, v)$ in the knowledge graph used in the query or answer options generation, we perform the following steps:

\begin{enumerate}
    \item \textbf{Collect Aliases and Text:} We gather aliases and the associated text paragraphs for both nodes $u$ and $v$.
    \item \textbf{Split into Sentences:} We split the text paragraphs of $u$ and $v$ into individual sentences using NLTK.
    \item \textbf{Identify Relevant Sentences:} We identify sentences that explicitly link the two nodes. A sentence from $u$'s text is considered relevant if it contains an alias of $v$, and vice versa. 
    \item \textbf{Discard Edges without Relevant Sentences:} If no relevant sentences are found for an edge, it is deemed unsupported and is discarded from the graph.
    \item \textbf{Prepend First Sentence:} To ensure the entity's primary name or common alias is included, we prepend the first sentence of each node's text to its list of relevant sentences.
\end{enumerate}

\subsubsection{Relevant sentences per Edge PrimeKG}
As we synthetically generate data for each edge in PrimeKG, we already have a sentence that corresponds to an edge in primeKG. So no further processing is needed.

\subsubsection{Trimming to Fit Context Length}

When constructing the final prompt for the LLM, we prioritize including the most relevant information within the model's context length limit. Therefore we need to trim the context according to the LLM's context limit.
We use the following procedure (detailed in Algorithm~\ref{alg:contruct_context}):
\begin{enumerate}
\item \textbf{Create Sentence Lists:} We create three lists of sentences:
\begin{itemize}
\item $S_{all}$: Contains all sentences from the text paragraphs of nodes involved in the query and answer options.
\item $S_{query}$: Contains all relevant sentences for the edges that constitute the query path.
\item $S_{options}$: Contains all relevant sentences for the edges used to generate the answer options.
\end{itemize}
\item \textbf{Construct the Final Context:}
\begin{enumerate}
\item We prioritize including all sentences from $S_{query}$ as they are directly related to the query.
\item Next, we add as many sentences from $S_{options}$ as possible, given the remaining context length limit.
\item Finally, we fill the remaining space with sentences from $S_{all}$ that have not been included yet, ensuring no sentence is repeated.
\end{enumerate}
\end{enumerate}
\begin{algorithm}
\label{alg:contruct_context}
\caption{Context Construction}
\begin{algorithmic}[1]
\STATE \textbf{Input:} $S_{all}$, $S_{query}$, $S_{options}$, $L_{max}$
\STATE \textbf{Output:} $C_{trimmed}$
\STATE $C \gets S_{query}$
\STATE \textbf{ASSERT} {TokenizedLength($C$) $\leq L_{max}$}
\STATE $S_{seen} \gets$ UniqueSet($C$)
\FOR{each $s$ in $S_{option}$+$S_{all}$}
\IF{$s \notin S_{seen}$ \textbf{and} TokenizedLength($C$ + $s$) $\leq L_{max}$}
\STATE $C \gets C$ + $s$
\STATE Add $s$ to $S_{seen}$
\ENDIF
\ENDFOR
\STATE \textbf{return} $C$ as $C_{trimmed}$
\end{algorithmic}
\end{algorithm}

\paragraph{\textbf{Vanilla Setting Note: }} In the vanilla setting, we only want to check simple knowledge comprehension abilities of LLMs, so if our DAG used to create a query has an edge type (alias), we ensure that the provided context trims any sentences which are in the relevant sentences of the same edge type has any edge type in the DAG. In \textbf{distractor} setting, there is no such constraint.

\needspace{40\baselineskip}
\subsection{Few-Shot Examples}~\label{sec:few_shot_examples}

To guide the LLM towards the desired response format and demonstrate the reasoning process, we include 2 few-shot examples in the prompt (common across all prompts to all models). These examples provide a clear illustration of how to approach the knowledge comprehension task.
We investigate the impact of varying number of few-shot examples on LLM performance in Appendix~\ref{app:ablation}.
We use the following few-shot examples for wikidata5m:

\begin{tcolorbox}[
    colback=lightblue,
    colframe=blue!78!black,
    fonttitle=\bfseries,
    title=Few Shot Examples,
]

\textbf{Common Context:} 
entity\_B is the son of entity\_A. entity\_E is the sister of entity\_A. entity\_B leads entity\_C. Entity\_D is a member of Entity\_C. Entity\_D is a friend of entity\_E. entity\_E has mother entity\_F who likes the services of entity\_C.

\vspace{0.5em}

\vspace{0.5em}

\textbf{Question 1:} entity\_A \entityrel{father of} \entityrel{leader of} $\rightarrow$ ?

\textbf{Options:} 1. entity\_F, 2. entity\_C, 3. entity\_D, 4. entity\_E, 5. entity\_B

\textbf{Answer:} 2. \correctanswer{entity\_C}

\textbf{Explanation:} entity\_A \entityrel{father of} entity\_B \entityrel{leader of} entity\_C

\textit{How to get answer:} Find who entity\_A is father of to get entity\_B, then find what B is the leader of to get entity\_C.

\vspace{0.5em}

\vspace{0.5em}

\textbf{Question 2:} entity\_B \entityrel{chief of} \entityrel{constitutes} \entityrel{companion of} $\rightarrow$ ?

\textbf{Options:} 1. entity\_F, 2. entity\_C, 3. entity\_D, 4. entity\_E, 5. entity\_A

\textbf{Answer:} 4. \correctanswer{entity\_E}

\textbf{Explanation:} entity\_B \entityrel{chief of} entity\_C \entityrel{constitutes} entity\_D \entityrel{companion of} entity\_E

\textit{How to get answer:} Find what entity\_B is the chief of to get entity\_C, find what entity\_C constitutes to get entity\_D, then find the companion of entity\_D to get entity\_E.

\end{tcolorbox}

We use the following few-shot examples for primeKG:

\begin{tcolorbox}[
    colback=lightblue,
    colframe=blue!78!black,
    fonttitle=\bfseries,
    title=Few Shot Examples,
]

\textbf{Common Context:} 
(drug) entity\_A has side effects: effect\_1, effect\_5. (disease) entity\_C has contraindication relation to (drug) entity\_E. (disease) entity\_C is indicated for (drug) entity\_A.
                (drug) entity\_D has side effects: effect\_2, effect\_3, effect\_4. (drug) entity\_D has indication for (disease) entity\_C. (drug) entity\_E has indication to (disease) entity\_B. 
                (gene/protein) entity\_F is carrier for (drug) entity\_E. (disease) entity\_B is indicated for (drug) entity\_G.

\vspace{0.5em}

\vspace{0.5em}

\textbf{Question 1:} Which drug \entityrel{has indication for} (disease) entity\_C and has \entityrel{side effect} effect\_5 ?

\textbf{Options:} 1. entity\_F,\\n 2. entity\_E,\\n 3. entity\_D,\\n 4. entity\_A,\\n 5. entity\_B

\textbf{Answer:} 2. \correctanswer{entity\_A}

\textbf{Explanation:} entity\_A has the side effects effect\_5 while it is the indication of entity\_C, so answer is entity\_A.

\textit{How to get answer:} Find the drugs entity\_C is indicated for, which gives us entity\_A and then check which of them has the side effect\_5.

\vspace{0.5em}

\vspace{0.5em}

\textbf{Question 2:} What drugs should be avoided in (disease) entity\_C but are indicated for (disease) entity\_B?

\textbf{Options:} 1. entity\_F,\\n 2. entity\_A,\\n 3. entity\_D,\\n 4. entity\_E,\\n 5. entity\_B

\textbf{Answer:} 4. \correctanswer{entity\_E}

\textbf{Explanation:} entity\_B is indicated for entity\_E (drug), and entity\_E has contraindication to entity\_C.

\textit{How to get answer:} find the drugs entity\_B is indicated for to get entity\_E, and entity\_G. then find the contraindications of entity\_C to get entity\_E.
The intersection of these two sets gives the answer, which in this case is entity\_E.

\end{tcolorbox}

\subsection{Final Prompt}\label{app:final_prompt}

The final prompt presented to the LLM is constructed using a template (same for wikidata5m and primeKG) that incorporates several key elements:

\componentheader{blue}{Trimmed Context [\ref{sec:context_trim}]:} The relevant context extracted and trimmed.

\componentheader{softgreen}{Query [\ref{sec:querygen}]:} The multi-hop query.

\componentheader{softorange}{Answer Options [\ref{sec:answer_options}]:} The generated answer options, including the correct answer and distractors.

\componentheader{purple}{Few-Shot Examples [\ref{sec:few_shot_examples}]:} A set of examples demonstrating the desired response format.

The prompt template is structured as follows:

\begin{tcolorbox}[
    colback=lightblue,
    colframe=blue!78!black,
    fonttitle=\bfseries,
    title=Prompt Template,
    enlarge right by=0.2cm
]
{\color{purple}\{few\_shot\_examples\}}

Actual Query:
Given Context:
{\color{blue}\{context\}}

Answer the question:
{\color{softgreen}\{query\}}

answer the question by selecting the correct answer from the following options:

{\color{softorange}\{options\}}

\vspace{0.5em}

The format for beginning your response is:

correct answer: $<option\_number>. \; <answer>, \; because <succinct \; reason>$

follow this exact format and only choose from the given options
\end{tcolorbox}

\textbf{Estimating the number of unique prompts: } We estimate a lower bound on the number of unique prompts that can be generated from the Wikidata5m Knowledge Graph (KG) by quantifying the potential unique queries within the graph. Each query can be formulated into multiple prompts through variations in answer presentation, thus making query count a conservative estimate. We analyzed the 50 subgraphs employed in our experiments. For each subgraph, we calculated the number of unique paths(upto the maximum path length hyperparameter, $\maxpathlen = 4$) and calculated the number of possible queries for each path using the number of aliases for each each entity and relation within a path. This analysis provides an estimate of the unique query generation capacity inherent in subgraphs in our KG.

The mean number of unique queries was $3.04 \times 10^{15}$ with a median of $1.24 \times 10^{15}$. The minimum and maximum observed values were $1.36 \times 10^{12}$ and $1.46 \times 10^{16}$, respectively.

Importantly, these figures conservatively estimate the number of unique prompts, as they only consider query variations and not the diversity introduced by different answer options. The actual number of unique prompts is likely significantly larger, making exhaustive enumeration of all possible generated prompts infeasible.

\subsection{Response Checker function}~\label{app:checker}
We implement a simple response checker function to evaluate the correctness of the model's answers. The function is defined in Algorithm~\ref{alg:response_checker}. We write a regular expression to account for trivial formatting errors like extra spaces, brackets, incorrect punctuation, etc.

\begin{algorithm}
\label{alg:response_checker}
\caption{Response Checker}
\begin{algorithmic}[1]
\STATE \textbf{Input:} $model\_answer$, $correct\_answer\_num$
\STATE \textbf{Output:} $is\_correct$

\STATE $model\_answer \gets$ LowerCase($model\_answer$)
\STATE $correct\_answer\_num \gets$ LowerCase(ToString($correct\_answer\_num$))
\STATE $pattern \gets$ SpecializedRegularExpression("correct answer: " + $correct\_answer\_num$)

\IF{RegexMatch($pattern$, $model\_answer$)}
    \STATE $is\_correct \gets 1$
\ELSE
    \STATE $is\_correct \gets 0$
\ENDIF

\STATE \textbf{return} $is\_correct$
\end{algorithmic}
\end{algorithm}

\newpage
\section{Ablations}\label{app:ablation}
\subsection{Few Shot Prompts}
We conduct an ablation study to examine the impact of varying the number of few-shot examples on Gemini-Flash's performance in the vanilla task setting. While our default configuration uses two few-shot examples, we extend this analysis to include up to five examples. Interestingly, we observe no significant variation in performance across these different few-shot configurations. The results are presented below in ~\ref{tab:certificates_fewshot}. 

\begin{table*}[!h]
    \caption{Certification results for LLMs in vanilla setting with different number of few-shot examples}
    \centering
    \begin{tabular}{@{}lrrr@{}}
        \toprule
        Model & Avg. lower bound & Avg. upper bound & Avg. accuracy \\
        \midrule
        Gemini-1.5-Flash 2Shot (Default) & $0.46 \pm 0.06$ & $0.58 \pm 0.06$ & $0.52 \pm 0.06$ \\
        Gemini-1.5-Flash 3Shot & $0.46 \pm 0.06$ & $0.58 \pm 0.06$ & $0.52 \pm 0.06$ \\
        Gemini-1.5-Flash 4Shot & $0.46 \pm 0.07$ & $0.58 \pm 0.07$ & $0.52 \pm 0.07$ \\
        Gemini-1.5-Flash 5Shot & $0.46 \pm 0.07$ & $0.58 \pm 0.07$ & $0.52 \pm 0.07$ \\
        \bottomrule
    \end{tabular}
    \label{tab:certificates_fewshot}
\end{table*}

\subsection{Distractor Distributions}\label{app:distractor_dists}

To assess the impact of distractor distribution on model performance, we implement three distinct distractor distribution strategies:
\begin{enumerate}
\item Tail-weighted: Linearly increasing weights towards the tail end of the path, prioritizing distractors near the answer node. This serves as our default setting.
\item Head-weighted: Linearly increasing weights towards the head of the path, emphasizing distractors near the query's starting point.
\item Uniform: Equal probability of selecting distractors from any position along the path.
\end{enumerate}

We observe no significant differences in either of the settings. The results are presented in ~\ref{tab:certificates_dists} below.
\begin{table*}[!h]
    \caption{Certification results for Gemini-Flash with different distractor distributions}
    \centering
    \begin{tabular}{@{}lrrr@{}}
        \toprule
        Model & Avg. lower bound & Avg. upper bound & Avg. accuracy \\
        \midrule
        Gemini-1.5-Flash Setting 1 (Default) & $0.42 \pm 0.10$ & $0.55 \pm 0.10$ & $0.48 \pm 0.10$ \\
        Gemini-1.5-Flash Setting 2 & $0.42 \pm 0.11$ & $0.55 \pm 0.11$ & $0.48 \pm 0.11$\\
        Gemini-1.5-Flash Setting 3 & $0.42 \pm 0.11$ & $0.55 \pm 0.11$ & $0.48 \pm 0.11$ \\
        \bottomrule
    \end{tabular}
    \label{tab:certificates_dists}
\end{table*}

\newpage
\subsection{Model Performances with varying Path Length}
Among our certificates, we have queries of various lengths. Here we study the effects on models behavior on queries with varying length by considering the number of hops they require to reason to answer the query(which is 1 less than the path length). To do so, we refer to the number of hops to answer a query as k where $1 \leq$ k $<\maxpathlen$.

\begin{figure}[!t]
    \centering
    \includegraphics[width=0.8\linewidth]{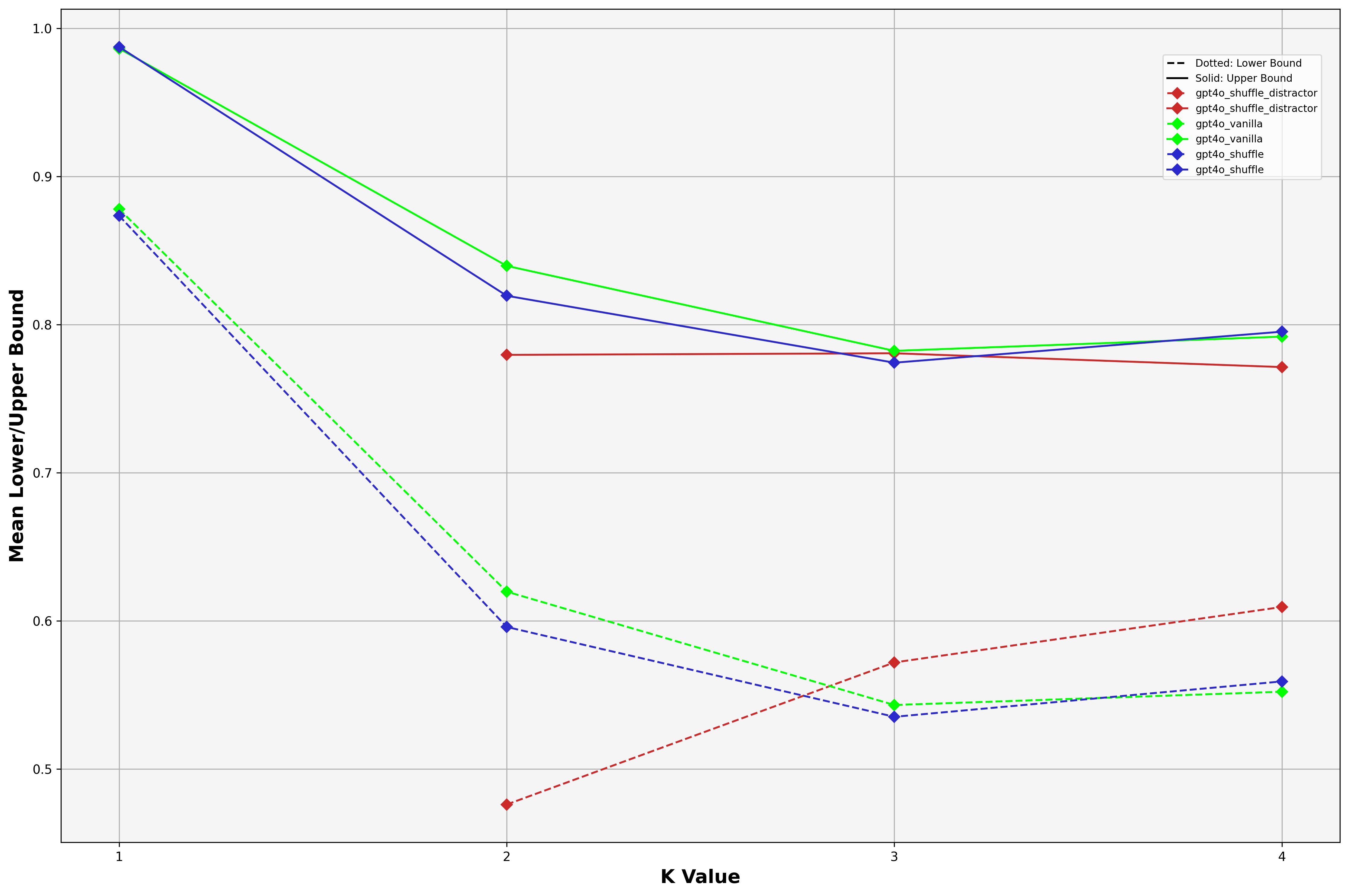}
    \caption{Variations in the bounds against the path lengths across various specifications.}
    \label{fig:k_plot_setts}
\end{figure}

\textbf{Varying Setting:} In figure ~\ref{fig:k_plot_setts} we show plots for various specifications for  the GPT4o model.

\begin{figure}[!h]
    \centering
    \includegraphics[width=0.8\linewidth]{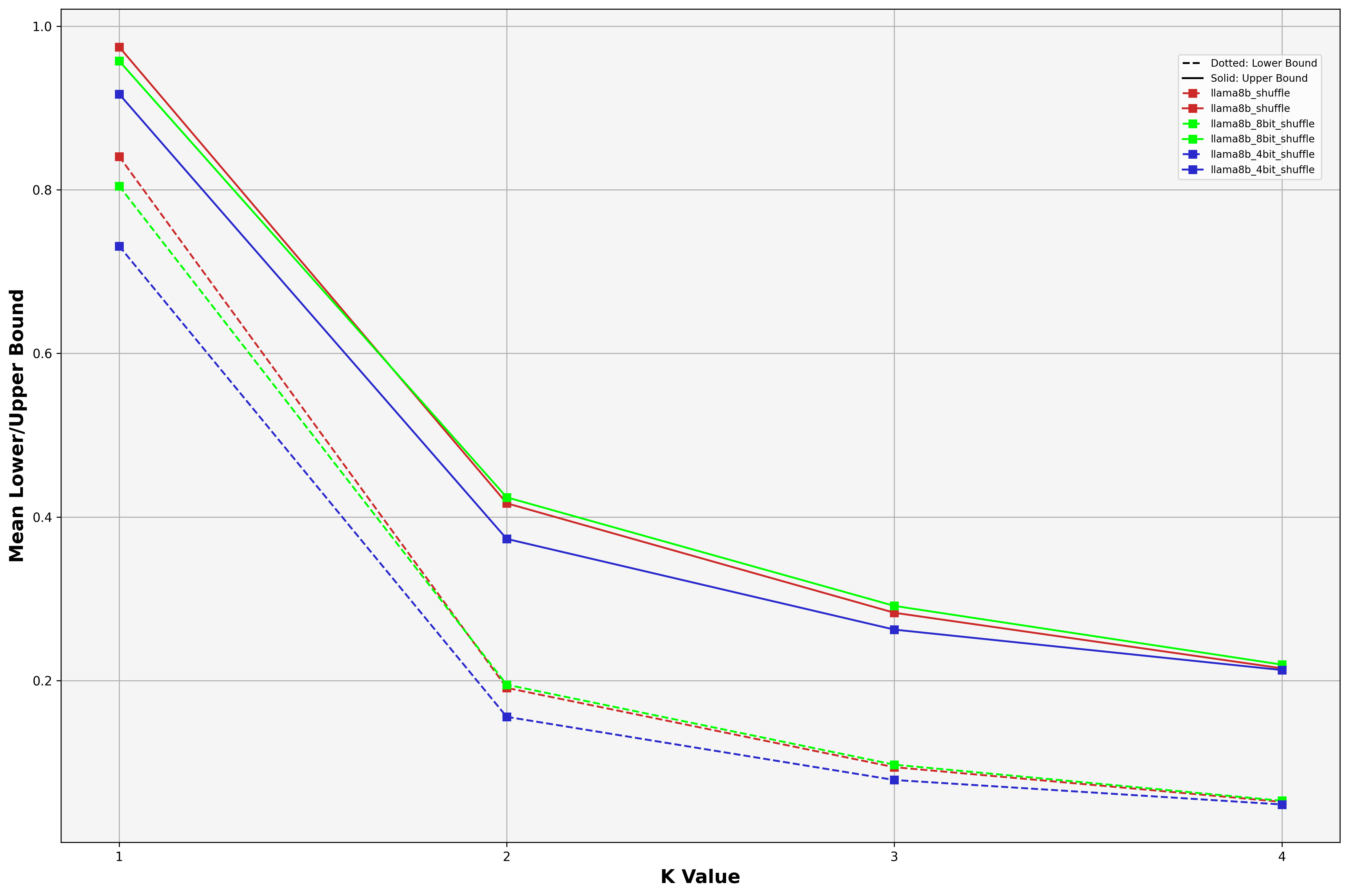}
    \caption{Variations in the bounds against the path lengths across various quantizations.}
    \label{fig:k_plot_quants}
\end{figure}

\textbf{Varying Quantization:} In figure ~\ref{fig:k_plot_quants} we show plots when the quantization is varied with the Llama3-8B model on the shuffle specification and their effects on performance.

\begin{figure}[!h]
    \centering
    \includegraphics[width=0.8\linewidth]{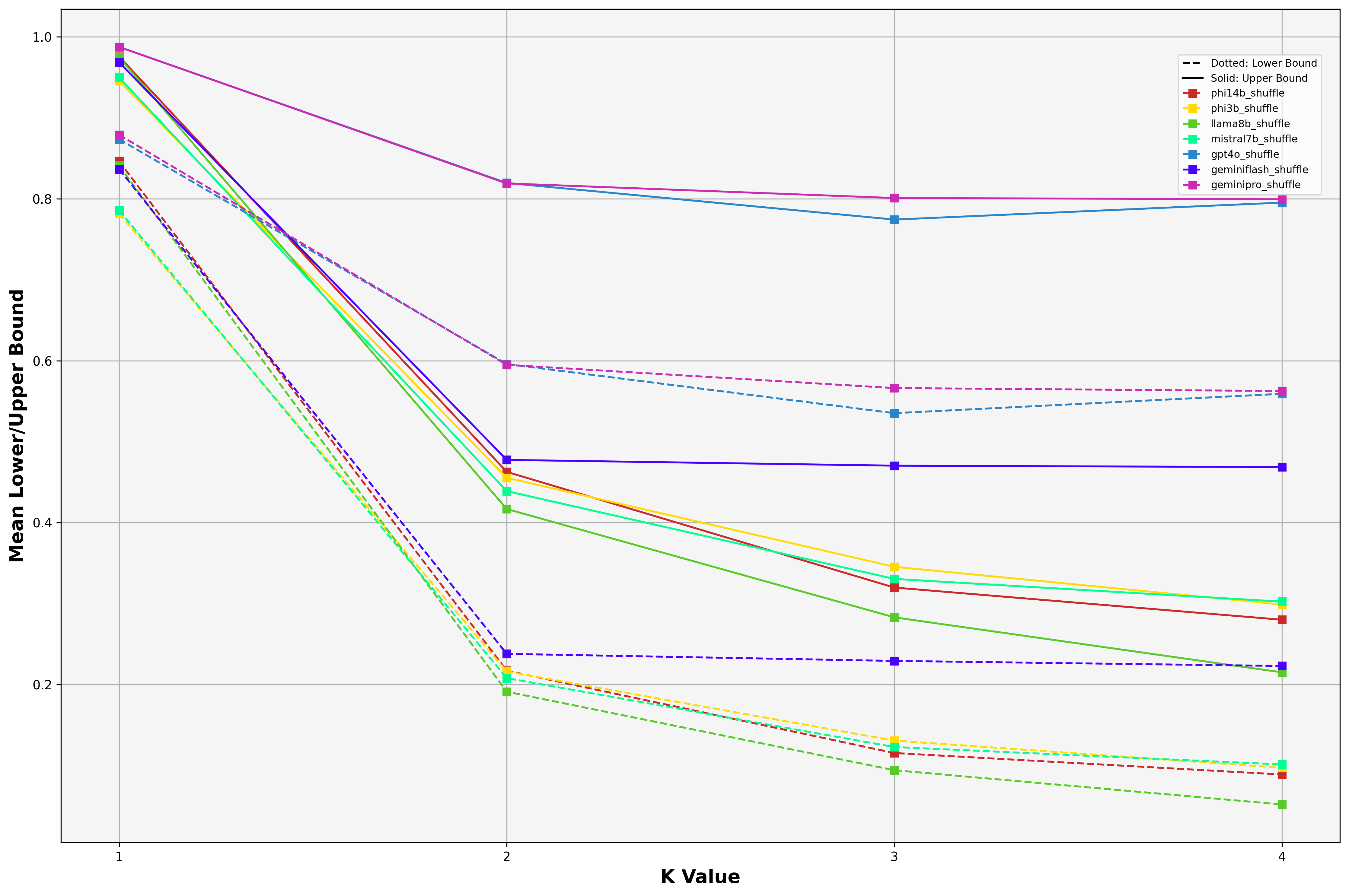}
    \caption{Variations in the bounds against the path lengths across various models in the shuffle setting.}
    \label{fig:k_plot_models}
\end{figure}

\textbf{Varying Models:} In fig ~\ref{fig:k_plot_models} we show plots for the shuffle specification and performance across the models(the open-source models use fp16 precision).
\newpage

In Figure ~\ref{fig:k_plot_setts}, we observe that the performance across settings converges as k increases and the distractor setting is most impactful on the performance for $k=2$.

In Figure ~\ref{fig:k_plot_quants}, we infer that as k increases the performance of the models' on the task converges across the different quantizations. We hypothesize this is due to the increasing complexity of the reasoning task.

In Figure~\ref{fig:k_plot_models}, we see that larger models (GPT-4o, Gemini-Pro) show less severe drop in performance compared to their smaller models. The figure shows that large models may have learnt to better apply 1-step reasoning for multiple steps when compared to their smaller counterparts.

\rebuttal{\subsection{Chain of Thought Prompting}\label{app:cot}
We also conduct an ablation on how Chain-of-Thought(COT) prompting can affect the performance of language models on the knowledge comprehension task. Specifically, we investigate the Phi-3 3B model (precision: float16) in the vanilla setting with COT prompting strategy. We augmented our standard few-shot examples ($\ref{sec:few_shot_examples}$) with COT steps and added structured reasoning guidance to the prompt template ($\ref{app:promptconst}$):

\begin{tcolorbox}[
    colback=blue!15!white,
    colframe=blue!75!black,
    fonttitle=\bfseries,
    title= COT additions to prompt template,
    enlarge right by=0.2cm
]

{Answer in the following the below format:

Let's solve this step by step:
1) Let's identify the starting point and path:
   - Start: [identify starting entity]
   - Path to follow: [break down the path components]

2) Let's follow the path:
   Starting from [entity]
   → [first relationship] → [next entity]
   → [next relationship] → [next entity]
   ... [continue as needed]

3) Verify our final destination reaches one of the given options

Therefore, the correct answer is: $<$option\_number$>$. $<$option\_text$>$}

\end{tcolorbox}

In the vanilla setting, adding COT prompting improved Phi-3 3B's performance, with the bounds increasing by 0.11 summarized in Table \ref{tab:cot_res}. While we acknowledge the potential benefits of COT, earlier experiments were limited due to the significantly increased computational cost (generating 5-8 times more tokens) and the expenses of COT, particularly with closed-source models as output tokens are much more expensive.

\begin{table}[!h]
    \caption{Certification results for Phi-3 3B with and without COT}
    \centering
    \begin{tabular}{@{}lrrr@{}}
        \toprule
        Prompting Strategy & Avg. lower bound & Avg. upper bound & Avg. accuracy \\
        \midrule
        No COT (Default) & $0.34 \pm 0.05$ & $0.46 \pm 0.06$ & $0.40 \pm 0.05$ \\
        COT & $0.45 \pm 0.08$ & $0.57 \pm 0.08$ & $0.51 \pm 0.08$\\
        \bottomrule
    \end{tabular}
    \label{tab:cot_res}
\end{table}

}

\section{Case Studies from certifying over Wikidata5m}\label{app:cs_wiki}
\begin{figure}[tb]
    \centering
    \includegraphics[width=\linewidth]{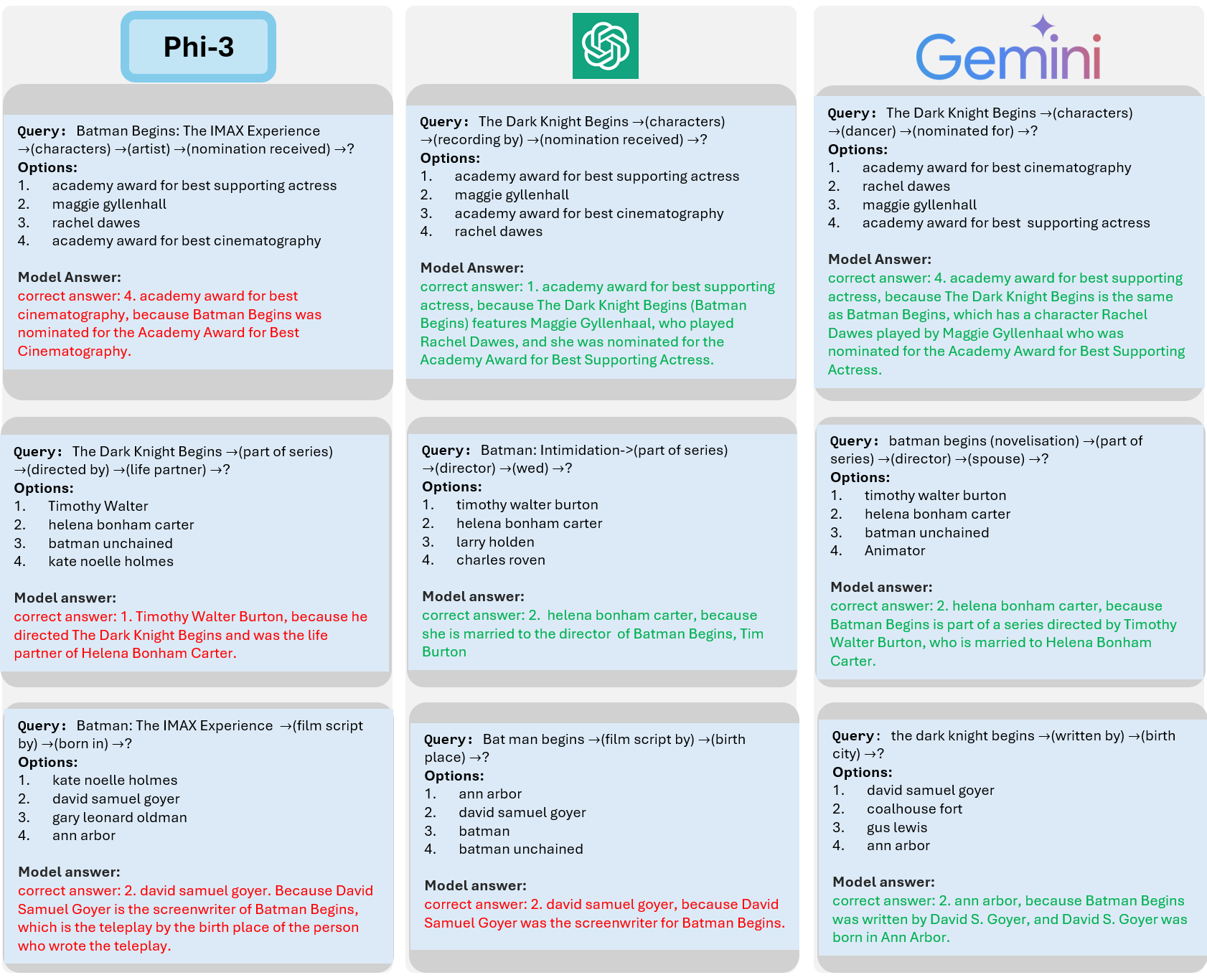}
    \caption{Qualitative analysis of samples used for certifying knowledge comprehension for Vanilla specifications on the Wikidata5m subgraph pivoted at the node for `Batman Begins' movie. The context provided in the prompts is not shown for brevity. Wrong model responses are colored red and correct ones are colored green. The samples are consistent with our results, wherein Phi-3 (3B) has lower certification bounds than GPT-4o's bounds, which are lower than those for Gemini-Pro.}
    \label{fig:cert_cs}
\end{figure}

\begin{figure}[tb]
    \centering
    \includegraphics[width=\linewidth]{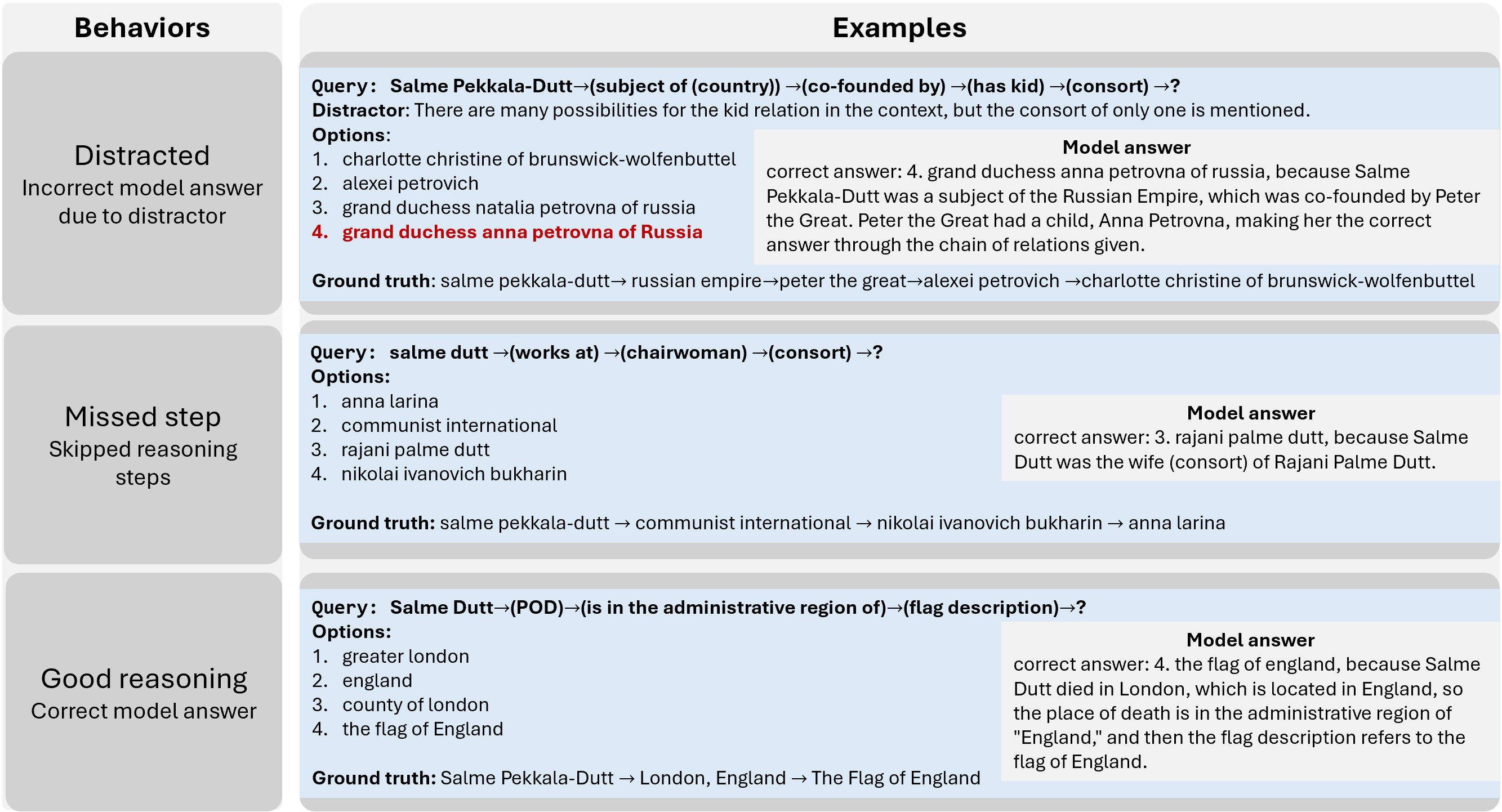}
    \caption{GPT-4o responses showing good reasoning, failures due to distractors or skipped reasoning}
    \label{fig:3_modes}
\end{figure}

We analyze the certification results, qualitatively. First, we show the responses of $3$ models in Figure~\ref{fig:cert_cs} --- Phi-3 (3B), GPT-4o, and Gemini, obtained when certifying them for the Vanilla specification defined over a subgraph pivoted at the node for `Batman Begins' movie. The samples reflect the certificates. 
Next, we identify and categorize prominent kinds of model responses. We frequently see the following failure modes --- \emph{distracted} and \emph{missed relation}. In the former, the model gets deviated from the query by following the distractor context in its prompt, resulting in an incorrect answer. In the latter, the model skips some reasoning steps needed for the final correct answer. In cases of \emph{good reasoning}, model accurately follows the query and gives the correct answer. Figure~\ref{fig:3_modes} presents examples of the aforementioned kinds of model responses for GPT-4o.


\end{document}